\documentclass[lettersize,journal]{IEEEtran}
\usepackage{amsmath,amsfonts}
\usepackage{algorithmic}
\usepackage{algorithm}
\usepackage{array}
\usepackage[caption=false,font=footnotesize,labelfont=rm,textfont=rm]{subfig}
\usepackage{textcomp}
\usepackage{stfloats}
\usepackage{url}
\usepackage{verbatim}
\usepackage{graphicx}
\usepackage{cite}
\usepackage{bm}
\usepackage{bbm}
\usepackage{multirow}
\usepackage{microtype}
\usepackage{booktabs}
\usepackage{array}
\usepackage{makecell}
\usepackage{microtype}
\usepackage{booktabs}
\usepackage{array}
\usepackage{makecell}
\usepackage{xcolor}
\usepackage{wrapfig}
\usepackage{soul}
\usepackage{thmtools}
\usepackage{thm-restate}
\usepackage{flexisym}
\usepackage{rotating}
\usepackage{colortbl} 
\usepackage{tikz,xcolor}
\usepackage[implicit=false]{hyperref}
\hypersetup{hidelinks,
	colorlinks=true,
	allcolors=black,
	pdfstartview=Fit,
	breaklinks=true}
\definecolor{lime}{HTML}{A6CE39}
\DeclareRobustCommand{\orcidicon}{
\begin{tikzpicture}
\draw[lime, fill=lime] (0,0)
circle[radius=0.16]
node[white]{{\fontfamily{qag}\selectfont \tiny \.{I}D}};
\end{tikzpicture}
\hspace{-2mm}
}
\foreach \x in {A, ..., Z}{%
\expandafter\xdef\csname orcid\x\endcsname{\noexpand\href{https://orcid.org/\csname orcidauthor\x\endcsname}{\noexpand\orcidicon}}
}

\begin{document}
\title{Attention Residual Fusion Network with Contrast  for Source-free Domain Adaptation}
\author{Renrong Shao\orcidA{},
        Wei Zhang\orcidB{},~\IEEEmembership{Member,~IEEE,}
        and Jun Wang\orcidC{},~\IEEEmembership{Member,~IEEE}
\thanks{
Received 17 December 2024; revised 22 April and 11 July 2025; accepted 19 October 2025. Date of current version 21 October 2025.
This work was supported by the National Key Research and Development Program of China under Grant No.2024YFE0212000. The associate editor coordinating the review of this manuscript and approving it for publication was Dr. Li Li. (Corresponding authors: Wei Zhang, Jun Wang)
  
Renrong Shao is with the Faculty of Military Health Services, Naval Medical University (Second Military Medical University), Shanghai 200433 China, and also with the School of Computer Science and Technology, East China Normal University, Shanghai 200263, China. (e-mail: roryshaw6613@smmu.edu.cn)

Wei Zhang and Jun Wang are with the School of Computer Science and Technology, East China Normal University, Shanghai 200263, China. 
Wei Zhang is also with KLATASDS-MOE.
(e-mail: zhangwei.ltt@gmail.com, wongjun@gmail.com).

}
}

\markboth{IEEE Transactions on Circuits and Systems for Video Technology, ~Vol.~, No.~, Nov.~2025}%
{Shell \MakeLowercase{\textit{et al.}}: A Sample Article Using IEEEtran.cls for IEEE Journals}




\maketitle
  
\begin{abstract}
   Source-free domain adaptation (SFDA) involves training a model on source domain and then applying it to a related target domain without access to the source data and labels during adaptation. The complexity of scene information and lack of the source domain make SFDA a difficult task. Recent studies have shown promising results, but many approaches to domain adaptation concentrate on domain shift and neglect the effects of negative transfer, which may impede enhancements of model performance during adaptation.
  In this paper, addressing this issue, we propose a novel framework of Attention Residual Fusion Network~(ARFNet) based on contrast learning for SFDA to alleviate negative transfer and domain shift during the progress of adaptation, in which attention residual fusion, global-local attention contrast, and dynamic centroid evaluation are exploited. 
  Concretely, the attention mechanism is first exploited to capture the discriminative region of the target object. Then, in each block, attention features are decomposed into spatial-wise and channel-wise attentions. The spatial-wise attentions are aggregated with original semantic features to achieve the cross-layer attention residual fusion progressively while the channel-wise attentions are exploited for self-distillation. 
  During adaptation progress, we contrast global and local representations to improve the perceptual capabilities of different categories, which enables the model to discriminate variations between inner-class and intra-class. 
  Finally, a dynamic centroid evaluation strategy is exploited to evaluate the trustworthy centroids and labels for self-supervised self-distillation, which aims to accurately approximate the center of the source domain and pseudo-labels to mitigate domain shift. 
  To validate the efficacy of our methods, we execute comprehensive experiments on five benchmarks of varying scales, i.e.,  Office-31, Office-Home, VisDA-C, DomainNet-126, Cub-Paintings.
  Experimental outcomes indicate that our method surpasses other techniques, attaining superior performance across SFDA benchmarks. \textit{Code is available at https://github.com/RoryShao/ARFNet.git.}
\end{abstract}


\section{Introduction}
\IEEEPARstart{D}{eep} learning has achieved great success in various tasks, such as classification~\cite{krizhevsky2012imagenet}, object detection~\cite{ren2015faster,liu2022domain}, and semantic segmentation~\cite{long2015fully}. 
However, the most successful fields of deep learning rely on huge amounts of labeled data to achieve a reliable level of generalization, which is extremely expensive in practical applications.
Besides, applying existing models trained on one scenario~(source domain) to other relevant scenarios~(target domain) usually fails to generalize well due to the discrepancy in the data distribution, which may be subject to two risks, namely,  domain drift and negative transfer. 
To solve this problem, unsupervised domain adaptation~(UDA) has been introduced to align the source and target domains by discrepancy minimization, without accessing the target label information either by learning a domain invariant feature representation~\cite{long2015learning,goodfellow2014generative, kumar2018co,zuo2022margin} or projecting the independent transformations to a common latent representation through adversarial distribution matching~\cite{tzeng2017adversarial, long2018conditional,saito2018maximum, tian2022unsupervised}. 

\begin{figure}[!tpb] 
  \centering 
    \includegraphics[width=0.9\linewidth]{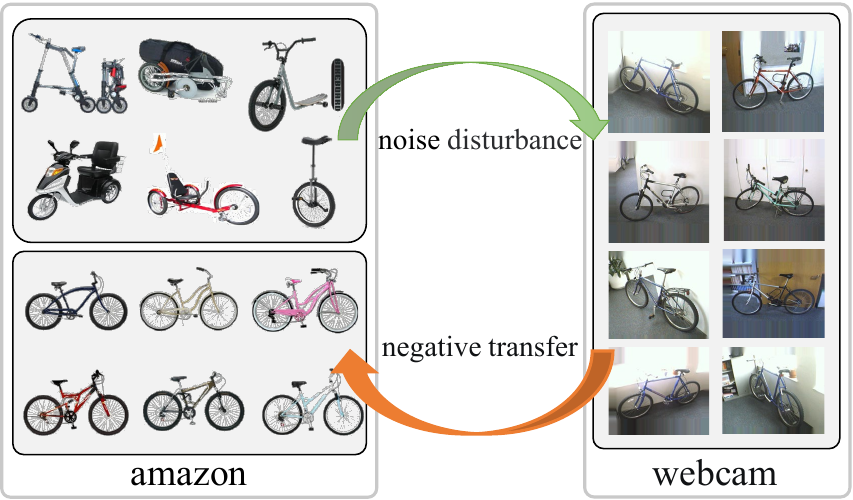}
    \caption{When samples from the real-world domain of Webcom are trained and adapted to fit the Amazon scenario within the Office31 dataset, two significant issues often arise, i.e., noise disturbance and negative transfer.}
    \label{fig:overview}  
  \vspace{-6mm}
\end{figure}
However, transfer learning is not always guaranteed, if the source and target domains are not sufficiently similar, transferring from such a weakly related source may hinder the performance in the target domain or even lead to negative transfer~\cite{wang2019characterizing,li2021semantic}.
The potential cause for negative transfer could be the unavoidable incorporation of cluttered background interference during the adaptation process of complex samples, thereby affecting the accurate matching of target samples~\cite{li2021semantic}.
As illustrated in Fig~\ref{fig:overview}, the white background samples of the Amazon in the Office31 dataset are easier to capture and discriminate, while those samples in the real world of Webcom are disturbed by background and lighting,  and thus relatively difficult to capture. 

To solve the negative transfer, some approaches propose to improve the concentration on distinct image regions~\cite{zhuo2017deep,li2020domain,li2021semantic,tian2022vdbda}.
For example, DUCDA~\cite{zhuo2017deep} designs a transfer mechanism for DA, which transfers the knowledge of discriminative patterns of source images to the target.
Differently, instead of exploring the space attention knowledge, DCAN~\cite{li2020domain} aims to excite distinct convolutional channels with a domain-conditioned channel attention mechanism. 
While SCAD~\cite{li2021semantic} proposes to achieve the semantic concentration for domain adaptation by leveraging dark knowledge.
Nevertheless, these studies are all conducted in a setting where case data is accessible.
In some specific scenarios, due to security privacy protection, and data transmission limitations, the model cannot access the source data directly. 
Although some prospective works~\cite{liang2020we, yang2021generalized, li2020model, xia2021adaptive} have been proposed for source-free domain adaptation~(SFDA), e.g., SHOT~\cite{liang2020we} proposes a self-supervised manner to achieve the alignment by pseudo-labels evaluation. 
G-SFDA~\cite{yang2021generalized} aims to cluster the target features with their similar neighbors.
3C-GAN~\cite{li2020model} proposes a collaborative class conditional generative adversarial net to bypass the dependence on the source data.
They did not consider the effect of negative transfer adequately in source-free scenarios. 

In this paper, we address this issue from a new perspective by a novel framework Attention Residual Fusion Network ~(ARFNet) based on contrast learning for SFDA to alleviate negative transfer and domain shift during the progress of adaptation, in which multilayer attention residual fusion~(MARF), global-local attention contrast~(GAC), and dynamic centroid evaluation~(DCE) are exploited. 
Concretely, we first introduce an attention feature extraction module in each block of the general backbone to capture the discriminative features from the object, which differs from the conventional semantic feature extraction~\cite{he2016deep}. 
Then, we decompose discriminative features into spatial-wise and channel-wise attentions.
To relieve the catastrophic forgetting dilemma and ensure the long-distance dependence of attention, we aggregate different-level attention features and fuse the cross-layer residual features of attention progressively.
During adaptation progress, we exploit GAC to contrast local attention features with memory-based global attention features which aim to improve the perceptual capabilities of different categories and enable the model to discriminate variations between intra-class and inter-class.
Finally, we exploit a DCE strategy to evaluate the trustworthy centroids and labels for self-supervised self-distillation~(SSD), which aims to accurately approximate the center of the source domain and pseudo-labels to mitigate domain drift. 

To verify our approaches, we conduct extensive experiments on five benchmarks of varying scales,  i.e., traditional Office-31, Office-Home, VisDA-C, DomainNet-126, and fine-grained Cub-Paintings.
Eventually, our approach can outperform the others and achieve superior performance among multiple domain adaptation benchmarks. Our contributions can be summarized as follows:
\begin{itemize}
   \item We propose a novel framework with MARF to capture and fuse the different level discriminative features of the target object, which aims to alleviate negative transfer.  

   \item  We propose to exploit memory-based GAC to contrast local and global attention features, which enable the model to discriminate variations between inner-class and intra-class.
   
   \item We proposed a DCE strategy to evaluate the credible centroids and labels for self-supervised self-distillation and mitigating domain drift. 
   
   \item Extensive experiments are conducted on five benchmark datasets to verify the effectiveness, which demonstrates our methods yield results comparable to the state-of-the-art~(SOTA) for unsupervised SFDA.
\end{itemize}

\section{Related Work}
\subsection{Source-free Domain Adaptation}
SFDA is the sub-field of Domain Adaptation, which tackles the issue of domain shift by aligning the feature distribution in the common projection feature space.
Recently, some prospective works have proposed to alleviate the issue of data privacy by SFDA~\cite{liang2020we, yang2021generalized, saito2018maximum, li2020model, xia2021adaptive, tian2022vdbda}.
Among them, SHOT~\cite{liang2020we} evaluates the pseudo-labels by constructing the centroid of coarse-grained clustering features, then aligns the source domain with the target domain by minimizing the information maximization loss. G-SFDA~\cite{yang2021generalized} evaluates the category consistency by the nearest neighborhood features.
MCD-DA~\cite{saito2018maximum} maximizes the discrepancy to detect target samples that are far from the support of the source and minimizes the discrepancy between the near support of generated target features.
3C-GAN~\cite{li2020model} develops a collaborative class conditional generative adversarial network for producing target-style training samples.
A$^{2}$Net~\cite{xia2021adaptive} adopts a soft-adversarial manner to adaptively distinguish source-similar target samples from source-dissimilar ones.
In this paper, we solve the negative transfer as well as domain shift of SFDA by proposing a new framework.

\subsection{Negative Transfer} 
Negative transfer (NT) is a pervasive issue in machine learning, where the act of transferring knowledge from a source dataset or model can adversely affect the performance of the target model~\cite{rosenstein2005transfer,pan2009survey, jiang2022transferability}. 
Initial research that identified negative transfer~\cite{rosenstein2005transfer} focused on rudimentary classifiers, such as hierarchical Naive Bayes. 
Subsequently, comparable negative impacts have been observed across diverse settings, encompassing multi-source transfer learning~\cite{duan2012exploiting}, imbalanced distributions~\cite{ge2014handling}, and partial transfer learning~\cite{cao2018partial}.
To counteract negative transfer, domain adaptation methods employ importance sampling or instance weighting strategies to prioritize relevant source data~\cite{wang2019characterizing,zhang2018importance}. 
Fine-tuning methods filter out harmful pre-trained knowledge by suppressing untransferable spectral components in the representation~\cite{chen2019catastrophic}. 
MTL methods use gradient surgery or task weighting to reduce the gradient
conflicts across tasks~\cite{yu2020gradient, liu2019loss}. 
Different from previous work, we propose to exploit attention to capture the discriminative features block-by-block progressively and fuse the features of each level, which is rare in this field yet will be beneficial for future research.

\subsection{Contrastive Learning for DA} 
Contrastive Learning has made remarkable progress in recent years and is widely exploited in unsupervised learning~\cite{oord2018representation,chen2020simple,he2020momentum,wang2020understanding}. 
The crucial to contrastive learning is mining effective embedding representations through deep models.
Normalized embeddings from the same category are then drawn closer together, while embeddings from different categories are mutually exclusive.
For instance, several representative contrastive learning paradigms have been developed specifically for categorization tasks, such as MoCo~\cite{he2020momentum}, SimCLR~\cite{chen2020simple}, and SupCL~\cite{khosla2020supervised}.
Recently, contrastive learning was mainly applied in the UDA setting~\cite{kang2019contrastive,park2020joint,kim2020cross}, where models have access to the source labels and(or) used models pre-trained on the ImageNet dataset as their backbone network. 
While CDA~\cite{thota2021contrastive} is based on contrastive learning, without having access to labeled data or pre-trained parameters, but instead leverages the vast amount of unlabeled source and target data to train an encoder from randomly initialized parameters. 
UC-SFDA~\cite{chen2023uc} establishes a novel SFDA framework based on uncertainty prediction and a neighborhood-guided evidence-based contrastive learning scheme.
In this paper, we utilize the traditional setting of contrastive learning for SFDA but innovate by proposing GAC to contrast the global and local views of samples in a naive way, which aims to improve the perceptual capabilities of different categories and enable the model to discriminate variations between inner-class and intra-class.
\subsection{Self Knowledge Distillation} 
Knowledge distillation~(KD) aims to transfer knowledge from a cumbersome large teacher model to a lightweight student model, which has been widely researched in recent years due to model overload and privacy protection. Since the original introduction was proposed in~\cite{hinton2015distilling}, many works extended the application of KD. For example, FitNet~\cite{romero2014fitnets} focuses on the intermediate layers' representations by exploiting regression to match the features of teacher and student. RKD~\cite{park2019relational} utilizes the distance-wise and angle-wise distillation losses that penalize structural differences in relations. 
Self-knowledge distillation~(SKD) or self-distillation~(SD) is a sub-field of KD, which aims to enhance efficiency and effectiveness in knowledge transferring. It is proposed to utilize knowledge from itself, without the involvement of extra networks~\cite{wang2021knowledge}. Commonly, there are three popular ways to construct an SD model. i.e., 1) data distortion-based self-distillation~\cite{lee2020self, xu2019data}, 2) use history information as a virtual teacher, 3) distilling across auxiliary heads~\cite{zhang2019your}. Our self-distillation is similar to the first, but unlike the previous works, our self-distillation is achieved by channel-wise attention, which can help the model improve the effectiveness of the target region acquisition. 

\begin{figure*}[t]
  \centering
   \includegraphics[width=\linewidth]{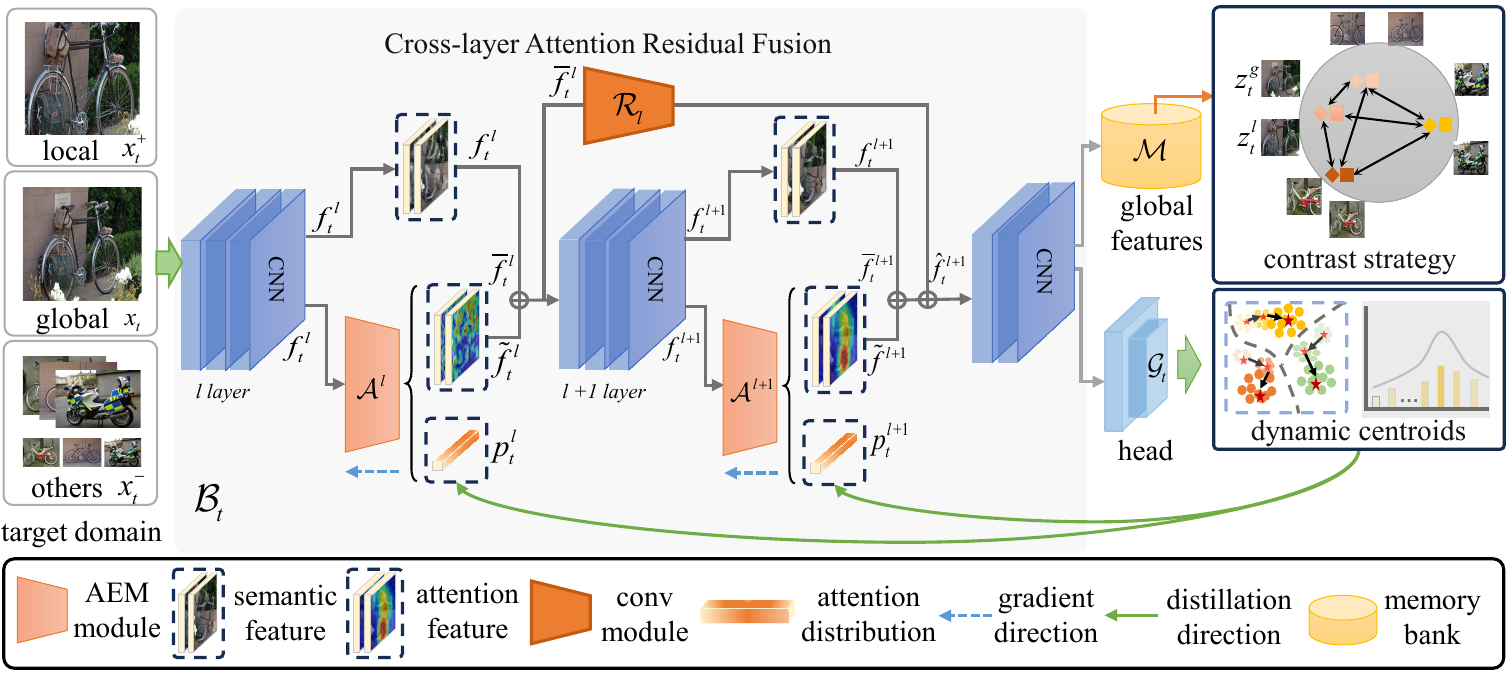}   
     \caption{The overall workflow of our proposed framework on the target domain. The components of trainable backbone $\mathcal{B}_{t}$ and frozen head $\mathcal{G}_{t}$, they are initialized by the well-trained source parameters of $\mathcal{B}_{s}$ and $\mathcal{G}_{s}$, respectively. The input is global views $x$ and corresponding local views $x^{+}$.
     In the backbone, an AEM module is introduced to each block, which can capture the attention of the feature. Then, attention and semantic features are fused to obtain fusion feature representations, which are used for MARF. Latent vectors $z$ are obtained from the $\mathcal{B}_{t}$ and projected to feature representations to execute GAC. In the output layer, the DCE is exploited to evaluate the trustworthy centroids and labels for self-supervised self-distillation learning. }
   \label{fig:framework}
   \vspace{-5mm}
\end{figure*}

 \section{Method}
 \subsection{Preliminaries}
 \noindent \textbf{Notation.} In this paper, we address the unsupervised SFDA task only with a pre-trained source model and without access to any source data.
 We consider image classification and denote the labeled source domain data with $n_{s}$ samples as $\mathcal{D}_{s} = { \left \{ (x_{s}^{i}, y_{s}^{i}) \right \} }^{n_{s}}_{i=1}$ where $x^{i}_{s} \in \mathcal{X}_{s}, y^{i}_{s} \in \mathcal{Y}_{s}$, and unlabeled target domain data with $n_{t}$ samples as $\mathcal{D}_{t} = { \left \{ x_{t}^{i}  \right \} }^{n_{t}}_{i=1}$, where $ x^{i}_{t} \in \mathcal{X}_{t} $. The goal of SFDA is to train a network of $\mathcal{G} \circ \mathcal{B}$ with the given source data to predict the labels $\left \{  y^{i}_{t} \right \}^{n_{t}}_{i=1}$ of unlabeled data in the target domain, where $\mathcal{B}(\cdot; \theta_{\mathcal{B}}): x \to z$ denotes the backbone of feature extractor, which obtain the latent vector representation $z$ from the raw images $x$, and $\mathcal{G}(\cdot; \theta_{\mathcal{G}}): z \to p$ denotes the linear classifier. 
 
 \noindent \textbf{Source domain model pretraining.}
 In our work, we first equip the backbone with the parameters pre-trained on ImageNet~\cite{deng2009imagenet} and later fine-tune it on the source domain, which aims to obtain a well pre-trained source classifier as follows:
 {\small
 \begin{align}
 \label{eq:source_ce}
 \mathcal{L}^{s} \left(\mathcal{X}_{s}, \mathcal{Y}_{s}\right)&=
 \mathbb{E}_{\left(x_{s}, y_{s}\right) \in \mathcal{X}_{s} \times \mathcal{Y}_{s}} \sum_{k=1}^{K}  \mathcal{L}_{ce}( \mathcal{G} \circ \mathcal{B} (x_{s}), y_{s}) \,, 
 \end{align}
 }
 
\noindent where $\mathcal{G} \circ \mathcal{B}$ denotes the whole network, the output of the network is a K-dimensional vector, and $\mathcal{L}_{ce}(\cdot,\cdot)$ is the cross-entropy loss. Then, we transfer the trained model to the target domain, solving the negative transfer and domain shift by MARF, GAC, and DCE, which are detailed in the following section. The overall workflow of our proposed framework on the target domain is indicated in Fig.\ref{fig:framework}.

 \subsection{Multilayer Attention Residual Fusion}
 For DA tasks, there are two main challenges: domain shift and negative transfer~\cite{pan2009survey, jiang2022transferability}. 
 Previous studies have primarily focused on and addressed domain shift, yet there has been limited research conducted specifically on negative transfer. Some research attempts to mitigate negative transfer from different aspects, such as importance sampling or instance weighting strategies~\cite{wang2019characterizing,zhang2018importance} or filtering out harmful pre-trained knowledge by suppressing untransferable spectral components in the representation~\cite{chen2019catastrophic}. However, they ignore the shortcomings of the architecture itself.
 The major shortcomings of traditional model architectures are that they cannot focus on discriminative foreground features and are susceptible to contextual influences~\cite{zhou2019discriminative}.
 Despite existing numerous  research~\cite{moon2017completely,hu2018squeeze,zhou2019discriminative} propose to exploit the attention mechanism to capture the discriminative region of the target object for different tasks, there is no such specific design for SFDA.
 Therefore, from the different aspects to fill this gap, we are the first propose to introducing MARF to improve the backbone of the model to eliminate the influence of this factor. We term this improved network as ARFNet.
 As shown in Fig.~\ref{fig:framework}, exploiting the ResNet family of networks as fundamental backbones, we divide the entire architecture into four blocks through the residual networks. Each block of backbone is denoted as $l$-th block, where $0 \leq l \leq 4$. An attention extract module~(AEM) $\mathcal{A}_{l}(\cdot;\theta_{\mathcal{A}_{l}}): f^{l}_{t} \rightarrow \tilde{f}_{t}^{l}, p^{l}_{t}$ is introduced in $l$-th block of the backbone $\mathcal{B}(\cdot; \theta_{\mathcal{B}})$, which aims to extract discriminative attention $\tilde{f}_{t}^{l}$ and channel-wise distribution $p^{l}_{t}$ from semantic features $f_{t}^{l}$ at different levels.
 Then, fusing the attention and semantic residual features $\bar{f}_{t}^{l}$ by the MARF block-by-block progressively, which aims to eliminate the pitfalls of long-distance decay of attention and aggregate different level attention features.
 In the process, we introduce a residual connection module $\mathcal{R}(\cdot; \theta_{\mathcal{R}}): \bar{f}^{l}_{t} \in \mathcal{R}^{C^{l}\times H \times W}\rightarrow \bar{f}^{l+1}_{t} \in \mathcal{R}^{C^{l+1} \times H \times W}$ to improve channel-wise dimensions and make neighboring blocks consistent.
    \begin{align}
    \begin{cases} 
      \tilde{f}_{t}^{l}, p^{l}_{t} &=\mathcal {A}^{l}(f^{l}_{t}; \theta_{\mathcal{A}^{l}}) \,,\\
      \bar{f}_{t}^{l} &= \tilde{f}_{t}^{l} \oplus f^{l}_{t} \,,  \\  
      \tilde{f}_{t}^{l+1}, p^{l+1}_{t}&=\mathcal {A}^{l+1}(f^{l+1};\theta_{\mathcal{A}^{l+1}}) \,, \\ 
      \bar{f}_{t}^{l+1} &= \tilde{f}_{t}^{l+1}
      \oplus f_{t}^{l+1} \,,  \\   
      \hat{f}_{t}^{l+1} &= \frac{1}{2} [ \mathcal {R}^{l}(\bar{f}_{t}^{l}; \theta_{\mathcal{R}^{l}} ) \oplus \bar{f}_{t}^{l+1} ] \,, \\
     \end{cases}   
  \end{align}
  
\noindent where the $l$ and  $l+1$ indicates the $l$-th block and the adjacent $l+1$ block of the backbone network, respectively. 
The $\mathcal{A}$ denotes the AEM, while $\mathcal{R}$ indicates the residual connection module, which can simply exploit Conv2D or Linear layer.
The sign $\oplus$ denotes the aggregation operation. The backbone with the block can extract the discriminative feature representations for domain adaptation. 
\subsection{Global and Local Attention Contrast} 
 Through the MARF proposed above, discriminative feature representations $\bar{f}_{t}$ of the attention and semantic features at different levels are obtained. Further, the backbone $\mathcal{B}(;\theta_{\mathcal{B}})$ extracts and projects the features to feature representations $z$ from output fusion features $\hat{f}_{t}$.
 Currently, contrastive learning has made remarkable progress in recent years and is widely exploited in unsupervised learning~\cite{oord2018representation,chen2020simple,he2020momentum,wang2020understanding}. Also, some research has introduced contrastive learning for DA tasks in a common manner~\cite{thota2021contrastive,huang2022category,zhang2022divide}.
 In contrastive  learning~\cite{chen2020simple,khosla2020supervised}, different pairs of positive and negative samples need to be constructed, such as data augmentation. 
 Different from the traditional approaches, to improve the perceptual capabilities of different categories, and enable the model to discriminate variations between inner-class
 and intra-class, we exploit GAC to contrast local and global attention features. For a target sample $x_{t} \in  \mathcal{X}_{t}$, we first construct a pair views of global and local as $(x^{g}_{t}, x^{l}_{t})$, where local view $x^{l}_{t}$ crops from the center of $x_{t}$ as the anchor and corresponding original sample $x^{g}_{t}$ as the positive view. Other instances are treated as negative samples.  
 Therefore, for all the target data $x^{i}_{t} \in \mathcal{X}_{t}$, where $i \in n_{t}$, the initialized model extract the pair views to achieve the feature representations as $\mathcal{B}((x^{g_{i}}_{t},x^{l_{i}}_{t});\theta_{\mathcal{B}}) \rightarrow (z^{g_{i}}_{t}, z^{l_{i}}_{t})$, as indicated in the Fig.~\ref{fig:framework}. 
 Besides, to contrast more diverse feature representations, we employ a memory-bank $\mathcal{M}(\cdot)$ to store all global-local feature representations $  z^{i}_{t} =  \left\{(z^{g_{i}}_{t}, z^{l_{i}}_{t}) \right\}, i \in n_{t}, z^{i}_{t} \in \mathcal{Z}_{t} $  as $\mathcal{M}( \mathcal{Z}_{t} )$ and update them by the mini-batch. 
 We exploit the contrastive manner to compute the discrepancy between global-local representations and other different instance representations by the following formula:
 
 \begin{align}
    \mathcal{L}_{gac} = - \mathbb{E}_{x^{i}_{t} \in \mathcal{X}_{t}} \left [ \mathrm{log}  \frac{ exp(z^{g_{i}}_{t} \cdot z^{l_{i}}_{t} / \tau)  }{\sum_{j=1}^{\mathrm{n_{t}}} \mathbbm{1}_{[i \ne j]}  exp(z^{l_{i}}_{t} \cdot z^{j}_{t} / \tau)} \right ]   \,,
    \label{eq:att_cl}
 \end{align}
 where $\mathbbm{1}_{[i \ne j]} \in \left \{0, 1 \right \} $ is an indicator function, $\cdot$ denotes the dot product, $\tau$ denotes the temperature parameter, $l_{i}$ denotes the index of local (anchor) representation, $g_{i}$ denotes the corresponding index of positive representation, and $j$ denotes the index of rest instance pair representation. For a batch of samples, the loss can decrease intra-class variation and increase inter-class variation at the same time. Therefore, it can be utilized to improve the perceptual capabilities of different categories.

\subsection{Dynamic Centroid Evaluation for SSD}
 Another challenge in domain adaptation is the issue of domain shift.
 For the source-driven setting, the crucial to domain adaptation is to find the invariant feature representations of the two domains. While in the source-free scenario, it is difficult to construct invariant features without the source domain. 
 Previous research has demonstrated that even though models trained in the source domain may not perform well in the target domain, they still distribute around certain clustering centers due to the innate invariance between the source and target domains~\cite{liang2020we,yang2021transformer}.
 Therefore, an effective and common approach is employed to construct clustering centers of features to align target samples~\cite{liang2021source}. Within this strategy, the evaluation of each sample's label is determined by calculating the distance between the feature and the centroid.
 However, due to the influence of noise samples, there are some biases in the centroids between the target and source, which may affect the evaluation of the centroids and impact the final model performance.
 Therefore, in this part, we proposed this DCE strategy, which aims to capture the centroids of each iteration. 
 To obtain the dynamic centroids, we first exploit the classifier $\mathcal{G}(\cdot; \theta_{\mathcal{G}}): z \rightarrow p$ to project the global feature representations of $z^{i}_{t}$ to logits $p^{i}_{t}$. The centroids for each category are evaluated by the clustering algorithm as ~\cite{caron2018deep,liang2020we}, however, we exploit the exponentially moving smoothing~(EMS) to dynamically evaluate each iteration of the overall progress as follows:
\begin{align}
\label{eq:centerid}
c_{k}^{i} & = \frac{\sum_{x_{t} \in \mathcal{X}_{t}} exp \left( z^{i}_{t} \cdot p^{i}_{t} \right) }{\sum_{x_{t} \in \mathcal{X}_{t}} exp \left(  p^{i}_{t} \right)} \,, \\   \notag
c_{k}^{i} & = \lambda c_{k}^{i-1} + (1-\lambda)c_{k}^{i}  \,,
\end{align}
where $exp$ denotes the exponential function, $\lambda$ represents the coefficient of movement, $k$ is category of samples. This function can evaluate the centroid of each cluster and find the nearest neighbor features by minimizing the distance with 
$\hat{y}_{t}^{i} = \arg \min _{k} D_{c}\left(z^{i}_{t}, c^{i}_{k}\right)$, 
where $D_{c}(\cdot, \cdot)$ indicates the cosine similarity measure function between feature representations and centroids. 
Then we compute the new target centroids based on pseudo-labels and minimize the distance to evaluate the pseudo-labels. Essentially, DCE strategy involves achieving label resmoothing through the EMS. This technique helps eliminate the randomness associated with centroid evaluation across multiple iterations thereby suppressing the noise of pseudo labels. 
Finally, target labels are obtained as:
\begin{align}
    \label{eq:pseudo_labels_set}
    \hat{y}_{t}= \mathrm{arg} \min_{k}( \sum_{x_{t} \in \mathcal{X}_{t}} \frac{ \mathbbm{1}\left(\hat{y}_{t}=k\right) z^{i}_{t}}{\mathbbm{1}\left(\hat{y}_{t}=k\right)}, c^{i}_{k})  \,,
\end{align}
where $\mathbbm{1}$ is the indicator function which means if and only if the $\hat{y}_{t}$ is equal to $k$, $z^{i}_{k}$ is obtained to compute the minimization distance with centerid $c^{i}_{k}$.
Finally, the self-supervised pseudo-labels $\hat{y}_{t}$ are obtained, it was generated by the DCE strategy in an unsupervised manner but can be used to align the target domain in a self-supervised manner. 

As mentioned above, through the dynamic centroids evaluation, we can obtain a pseudo-label with high confidence, which can be utilized for self-supervised and distillation training of the target domain.
{\small
    \begin{align}
      \label{eq:dynamic_label}
     \mathcal{L}_{ssd} & = \mathcal{L}_{ce} + \mathcal{L}_{kd} \\ \notag
     & = - \frac{1}{K} \sum_{k=1}^{K} \mathbbm{1}_{\left[k=\hat{y}_{t}\right]} \log \delta_{k}\left( p_{t} \right) + \sum_{l=1}^{L} \delta_{l}(p^{l}_{t})\log\delta(\frac{p^{l}_{t}}{p_{t}}) \,, \notag
    \end{align}
 }

\noindent where $\delta_{k}$ is the softmax, $l \in L$ is the layer of the model. We distill the dark knowledge from the output for channel-wise attention of each intermediate layer.

\subsection{Objective Loss}  
 Besides, to avoid model collapse in training and improve the quality of
 pseudo labels, the information maximization loss (indicated as $\mathcal{L}_{im}$) is exploited to maximize the distribution in target adaptation as 
\begin{align}
 \mathcal{L}_{im} & = 
 -\frac{1}{n_{t}} \sum_{i=1}^{n_{t}}\left\langle \delta(p_{t}^{i}), \log\delta(p_{t}^{i}) \right\rangle + \sum_{k=1}^{K} \hat{p}^{k}_{t} \log \hat{p}^{k}_{t},
\end{align}
 where $\delta(p^{i}_{t})$ is the softmax predicition of target sample $x^{i}_{t}$, $\hat{p}^{k}_{t}$ is the $k$-th element of $\hat{p}_{t} = \frac{1}{n_{t}} \sum_{i=1}^{n_{t}} \delta(p_{t}^{i}) $ and  $\left\langle \cdot, \cdot \right\rangle$ is the inner product operation. 
 
 To this end, we define the final total loss as the following:
 \begin{align}
  \label{eq:final_loss}
   \mathcal{L}^{t}_{total} =  \mathcal{L}_{im} + \alpha \mathcal{L}_{ssd} +  \beta  \mathcal{L}_{glac}  \,,
 \end{align}
 where $\alpha > 0$ and $\beta >0$  are the hyperparameters to control the relative influence of the corresponding terms for training. 

\begin{algorithm}[ht]
\caption{SFDA Pipeline of ARFNet}
\label{alg:algorithm}
\textbf{Input}: A pre-trained model $\mathcal{G}_{s} \circ \mathcal{B}_{s}$ on the source domain, target domain $\mathcal{X}_{t}$, hyperparameters $\alpha$, $\beta$, $\lambda$, and $\tau$, iteration $T$. \\
\textbf{Output}: Trained model for target domain $ \mathcal{G}_{t} \circ \mathcal{B}_{t} $.\\
\textbf{Initializtion}: Initialize $\mathcal{B}_{t}$ with parameters pre-trained on source domain, and freeze the classifier layer $\mathcal{G}_{t}$.

\begin{algorithmic}[1]
\FOR{\textit{epoch = 1} to \textit{T}}
\FOR{\textit{i = 1} to $n_{t}$}
\STATE Extract local feature representation and global attention features $(z^{g_{i}}_{t}, z^{l_{i}}_{t})$.
\STATE Store the pairs of global and local views to $\mathcal{M}_{(\cdot,\cdot)}$
\STATE Compare and update the representations of local feature and global feature.

\STATE Evaluate the dynamic pseudo-labels by Eq.~\ref{eq:dynamic_label}. \\
\STATE Align the target to source by SSD with Eq.~\ref{eq:final_loss}\\
\ENDFOR
\ENDFOR
\end{algorithmic}
\end{algorithm}

\section{Experiments}
\subsection{Datasets and Implementation Details}
\label{sec:detail}

\subsubsection{\textbf{Traditional and Fine-grained Datasets}} 
\noindent\textbf{Office-31}~\cite{saenko2010adapting} is a standard small-sized DA benchmark, which contains 4,652 images with 31 classes from three domains (Amazon (\textbf{A}), DSLR (\textbf{D}), and Webcam (\textbf{W})). 
\textbf{Office-Home}~\cite{venkateswara2017deep} is a challenging medium-sized benchmark, which consists of 15,500 images with 65 classes from four domains (Artistic images (\textbf{Ar}), Clip Art (\textbf{Cl}), Product images (\textbf{Pr}), and Real-World images (\textbf{Rw})). 
\textbf{VisDA-C}~\cite{peng2017visda} is a challenging large-scale \textbf{Sy}nthetic-to-\textbf{Re}al (\textbf{Sy}$\rightarrow$\textbf{Re}) dataset that focuses on the 12-class object recognition task. The source domain contains 152 thousand synthetic images generated by rendering 3D models while the target domain has 55 thousand real object images sampled from Microsoft COCO.
\textbf{DomainNet-126}~\cite{peng2019moment} 
is another large-scale dataset. As a subset of DomainNet containing 600k images of 345 classes from 6 domains of different image styles, which has 145k images from 126 classes, sampled from 4 domains, Clipart (\textbf{C}), Painting (\textbf{P}), Real (\textbf{R}), Sketch (\textbf{S}), as~\cite{saito2019semi} identify severe noisy labels in the dataset.
\textbf{CUB-Paintings} consists of two domains, i.e., CUB-200-2011(\textbf{C}) \cite{cub} and CUB-200-Paintings(\textbf{P}) \cite{PAN}, with significant domain shifts. Both include four-level granularity, encompassing 14 orders, 38 families, 122 genera, and 200 species. The former \textbf{C} has 11,788 images of real-world bird species, while the latter \textbf{P} is a collection of 3,047 images, including watercolors, oil paintings, pencil drawings, stamps, and cartoons.

\subsubsection{Implementation Details}
{Our methodology employs the ARFNet-50, ARFNet-101, ViT-B/16, and CLIP model with ResNet and ViT-B/32 architectures as its foundational backbones, which are systematically reconstructed from the ResNet-50, ResNet-101, and ViT-based frameworks.}
{As previously stated, we have incorporated an attention module into each block of the model and denatured the features to input them into the AEM model.} Building upon this modification, we have implemented a feature aggregation mechanism coupled with a cross-layer residual fusion strategy.
The model loads the parameters of ImageNet-1k to fine-tune the training on the source domain, with images resized to $224 \times 224$ pixels.
The learning rate for Office-31 and Office-Home is set to 3e-2, while for VisDA-C and DomainNet-126 is 1e-3. The training comprises 100 epochs and is optimized using stochastic gradient descent (SGD) with a momentum of 0.9 and a weight decay of 0.001. The default batch size is 64. The trade-off parameters $\alpha$ and $\beta$ are set as 1.0, 0.5,  while the other two parameters $\tau$, and $\lambda$ are set as 0.07, and 0.1, respectively. All experiments are conducted with PyTorch on NVIDIA A800 GPUs.
\renewcommand\arraystretch{1.1} 
\newcolumntype{Y}{>{\centering\arraybackslash}X}
\begin{table}[!htbp]
  \centering
  \vspace{-4mm}
  \caption{The \textbf{Left} are modules of channel-wise attention. The \textbf{Right} are modules of spatial-wise attention.}
  \resizebox{\linewidth}{!}{
    \begin{tabular}{c|c}
    
    \toprule

    Channel-wise Attention Modules  &  Spatial-wise Attention Modules \\

    \midrule

    [$B, C, H, W$] $ \rightarrow $ [$B, C, 1, 1$], & [$B$, $C$, $\cdot$, $\cdot$ ]$\rightarrow$ [$B$, $C/r$, $\cdot$, $\cdot$],  Conv(1 $\times$ 1)  \\
     AvgPool2d &  \\
    \midrule

    [$B, C, 1, 1$] $ \rightarrow $ [$B, C$], & [$B$, $C/r$, $\cdot$ , $\cdot$ ]  $\rightarrow$ [$B$, $2C/r$,  $\cdot$ , $\cdot$ ],  Conv($3\times3$), \\
    Reshape &  BN, ReLU, Maxpool  \\
        
    \midrule

    [$B, C$] $ \rightarrow $ [$B, C/r$], 
     & [$B$, $2C/r$, $\cdot$, $\cdot$ ]  $\rightarrow$ [$B$,  $4C/r$, $\cdot$, $\cdot$],  Conv($3\times3$),    \\
     Linear, ReLU  &  BN, ReLU \\
    
    \midrule

    [$B, C/r$]  $ \rightarrow $ [$B, C$], 
    & [$B$, $4C/r$, $\cdot$, $\cdot$ ]  $\rightarrow$ [$B$, $2C/r$, $\cdot$, $\cdot$] , Decov($3\times3$), \\
    Linear, Sigmod &  BN, ReLU, MaxUnpool \\
    
    \midrule

    [$B, C$] $ \rightarrow $ [$B, C, 1, 1$], 
    & [$B$, $2C/r$, $\cdot$, $\cdot$] $\rightarrow$ [$B$, $C/r$, $\cdot$, $\cdot$], Decov($3\times3$), \\
    Reshape  & BN, ReLU \\

    \midrule

    [$B, C, H, W$] $*$ [$B, C, 1, 1$],  [$B, C$]
    & [$B$, $C/r$, $\cdot$, $\cdot$]  $\rightarrow$  [$B$, $C$, $\cdot$, $\cdot$],  Conv($1\times1$), SoftMax \\

    \bottomrule

    \multicolumn{2}{l}{\small * $B$, $C$, $H$, and $W$ are the batch size, channel, height, and weight, while $\mathrm{r}$ is scale scalar.}
    \end{tabular}%
    }
    \label{tab:attention_modules}
\end{table}%

\subsubsection{Details of AEM}
{In ARFNet, we introduce the AEM module, which serves to extract attention from semantic features, ensuring that the model eliminates background interference and thus suppresses the generation of negative migration.}
Due to the task design, we need to decompose the attention into the spatial-wise and channel-wise.
Therefore, we follow the design of the SCA-CNN \cite{chen2017sca}, but do not recombine the decomposed attention, making it something different from the original to adapt our task.
The intricate architectural layout is delineated in the following Tab.~\ref{tab:attention_modules}, wherein the modular configuration of the Channel Attention mechanism is delineated on the left-hand side, contrasted with the modular design of the Spatial Attention mechanism on the right-hand side. 
{For channel-wise attention, the attention dimension is [B, C], where B represents the batch size and C denotes the channel corresponding to the number of categories. This setup can be utilized for self-distillation.}
For spatial-wise attention, the dimensions of input and output are the same, which are used for fusion with semantic features. {It is worth noting that this module is a plug-and-play component that can be easily plugged into different models.}

\begin{table}[htbp]
  \centering
  \caption{Accuracy~(\%) of different methods based on ResNet-50 for unsupervised SFDA in Office-31 dataset.}
  \resizebox{\linewidth}{!}{
    \begin{tabular}[b]{ll|c|ccccccc}
    \toprule
    Method & Venue & S.F. & A $\rightarrow$ D   &  A$\rightarrow$ W  & D$\rightarrow$ A   & D $\rightarrow$ W   & W $\rightarrow$A   & W $\rightarrow$ D   &  Avg. \\
    \midrule
    Source-Only~\cite{he2016deep}& CVPR16 & $\checkmark $ & 80.8 & 76.9 & 60.3 & 95.3 & 63.6 & 98.7 & 79.3 \\
    SHOT~\cite{liang2020we}& ICML20  & $\checkmark $   & 94.0  & 90.1  & 74.7  & 98.4  & 74.3  & 99.9  & 88.6 \\
    3C-GAN~\cite{li2020model}& CVPR20   & $\checkmark $& 92.7  & 93.7  & 75.3  & 98.5  & 77.8  & 99.8  & 89.6 \\
    SFDA~\cite{kim2021domain}& AI21  & $\checkmark $   & 92.2  & 91.1  & 71.0  & 98.2  & 71.2  & 99.5  & 87.2 \\
    A$^2$Net~\cite{xia2021adaptive}& ICLR21  & $\checkmark $ & 94.5 & 94.0 & 76.7 & \textbf{99.2} & 76.1 & \textbf{100.0} & 90.1 \\
    SFDA-DE~\cite{ding2022source}&  CVPR22  & $\checkmark $ & 96.0 & 94.2 & 76.6 & 98.5 & 75.5 & 99.8 & 90.1 \\
    PLUE~\cite{Litrico_2023}& CVPR23  &  $\checkmark $ & 89.2 & 88.4 & 72.8 & 97.1 & 69.6 & 97.9 & 85.8 \\
    {C-SFDA~\cite{Karim2023sfda}} & {CVPR23} &  {$\checkmark $} & {96.2} & {93.9} & {77.3} & {98.8}  & {77.9}  & {99.7} & {90.5} \\
    {ASOGE~\cite{cui2024adversarial}} & {TCSVT24} & {$\checkmark$} & {95.6} & {94.1} & {74.3} & {98.1}  & {74.2} &  {99.7} & {89.3} \\
    TPDS~\cite{tang2024source}& IJCV24  & $\checkmark $ & \textbf{97.1} & \textbf{94.5} & 75.7 & 98.7  & 75.5 & 99.8 & 90.2 \\
    {Improved SFDA~\cite{Mitsuzumi2024Understanding}} & {CVPR24} & {$\checkmark $} & {95.3} & {94.2} & {76.4} & {98.3} & {77.5} &  {99.9} & {90.3} \\
    \textbf{ARFNet}~(\textbf{Ours})& --  & $\checkmark $ & 96.6 &	93.6 &	\textbf{77.7} &	98.6 &	\textbf{78.3} &	99.8 &	\textbf{90.8}  \\
    \bottomrule
    \end{tabular}%
}
  \label{tab:office-31}%
  \vspace{-6mm}
\end{table}%

\subsection{Baselines and Comparison methods}
To further validate the effectiveness of our method on target domains, we compare the proposed approach with the SOTA methods mainly under the closed-set unsupervised SFDA.
For the closed-set DA, the compared methods include
source-free~(\textbf{S.F.}) methods of ResNet-50 and ResNet-101 backbones:
SFDA~\cite{kim2021domain},
SHOT~\cite{liang2020we},
3C-GAN~\cite{li2020model},
A$^2$Net~\cite{xia2021adaptive},
G-SFDA~\cite{yang2021generalized},
SFDA-DE~\cite{ding2022source},
DaC~\cite{zhang2022divide},
GKD~\cite{tang2021model},
NRC~\cite{yang2021nrc},
AdaCon~\cite{chen2022contrastive},
CoWA~\cite{lee2022confidence},
PLUE~\cite{Litrico_2023},
{C-SFDA~\cite{Karim2023sfda}},
{ASOGE~\cite{cui2024adversarial}},
TPDS~\cite{tang2024source},
{Improved SFDA~\cite{Mitsuzumi2024Understanding}};
We introduce our enhanced backbone, ARFNet-50, and rigorously compare its performance against the ResNet-50 backbone using small-sized and medium-sized datasets, specifically Office-31 and Office-Home. To ensure a fair assessment, we also evaluate both ARFNet-50 and ARFNet-101 alongside ResNet-50 and ResNet-101 on larger-scale datasets, namely DomainNet-126 and VisDA-C. This comprehensive analysis highlights the superior capabilities of our proposed architecture in diverse settings.
For each benchmark, we exploit the information maximization~(IM) loss $\mathcal{L}_{im}$ as our baseline.

\subsubsection{Results of Office-31}
The experimental results in Tab.~\ref{tab:office-31} reveal that directly exploiting the ResNet-50 backbone, trained on the source domain, for adaptation only obtains a result of 79.3\%.
For the ResNet-50 approaches in source-free settings, such as SHOT, A$^2$Net, SFDA-DE, and TransDA, the performance of these approaches can achieve around 90\%, which demonstrates that SFDA can achieve good performance even in a scenario without source data.
{Currently, the SOTA performance of ResNet-50 backbone on Office-31 is achieved by C-SFDA, while Improved SFDA achieves suboptimal performance.}
The last row is our approach, which significantly outperforms previously published SOTA approaches, advancing the average accuracy from {90.5\%~(C-SFDA)} to 90.8\%. 
Specifically, ARFNet outperforms other approaches primarily on the two specific tasks~(i.e., D $\rightarrow$ A and W $\rightarrow$ A), demonstrating the overall effectiveness of our method through the average score.

\begin{table*}[!htbp]
  \centering
  \caption{Accuracy (\%) of different methods based on ResNet-50 for unsupervised SFDA on the Office-Home dataset.}
  \resizebox{\linewidth}{!}{
    \begin{tabular}{ll|c|ccccccccccccc}
    \toprule
    \multicolumn{1}{c}{Method}& Venue & S.F. & Ar$\rightarrow$Cl &  Ar$\rightarrow$Pr &  Ar$\rightarrow$Rw  &  Cl$\rightarrow$Ar &  Cl$\rightarrow$Pr &  Cl$\rightarrow$Rw  & Pr$\rightarrow$Ar  & Pr$\rightarrow$Cl &  Pr$\rightarrow$Rw  & Rw$\rightarrow$Ar  & Rw$\rightarrow$Cl &  Rw$\rightarrow$Pr  & Avg. \\
    \midrule
    Source-only~\cite{he2016deep} & CVPR16 & $\checkmark$ & 44.5 & 67.6 & 74.7 & 52.5 & 62.8 & 64.9 & 53.1 & 40.5 & 73.3 & 65.3 & 45.1 & 77.9 & 60.2 \\
    G-SFDA~\cite{yang2021generalized}& ICCV21 & $\checkmark$ & 57.9 & 78.6 & 81.0 & 66.7 & 77.2 & 77.2 & 65.6 & 56.0 &  82.2 & 72.0 & 57.8 & 83.4 & 71.3 \\
    SHOT~\cite{liang2020we}& ICML20 & $\checkmark$ & 57.1  & 78.1  & 81.5  & 68.0  & 78.2  & 78.1  & 67.4  & 54.9  & 82.2  & 73.3  & 58.8  & 84.3  & 71.8 \\
    SFDA~\cite{kim2021domain}& AI21 & $\checkmark$& 48.4& 73.4& 76.9& 64.3& 69.8& 71.7& 62.7& 45.3& 76.6& 69.8& 50.5& 79.0& 65.7 \\
    A$^2$Net~\cite{xia2021adaptive}& ICCV21  & $\checkmark$& 58.4 & 79.0 & 82.4 & 67.5 & 79.3 & 78.9 & 68.0 & 56.2 & 82.9 & 74.1 & 60.5 & 85.0 & 72.8 \\
    DaC~\cite{zhang2022divide}& NIPS22  & $\checkmark$ & 59.1 & 79.5 & 81.2 & 69.3 & 78.9 & 79.2 & 67.4 & 56.4 & 82.4 & 74.0 & \textbf{61.4} & 84.4 & 72.8 \\
    SFDA-DE~\cite{ding2022source}& CVPR22  & $\checkmark$ & 59.7 & 79.5 & 82.4 & 69.7 & 78.6 & 79.2 & 66.1 & 57.2 & 82.6 & 73.9 & 60.8 & 85.5 & 72.9 \\ 
    PLUE~\cite{Litrico_2023} &  CVPR23 & $\checkmark$ &  49.1 & 73.5 & 78.2 & 62.9 & 73.5 & 74.5 & 62.2 & 48.3 & 78.6 & 68.6 & 51.8 & 81.5 & 66.9 \\
    C-SFDA ~\cite{Karim2023sfda} & CVPR23  & $\checkmark$  & 60.3 & 80.2 & \textbf{82.9} & 69.3 & \textbf{80.1} & 78.8 & 67.3 & 58.1 & \textbf{83.4} & 73.6 & 61.3 & 86.3 & 73.5 \\
    ASOGE~\cite{cui2024adversarial} & TCSVT24 & $\checkmark$ & 59.1 & 78.4 & 81.0 & 67.7 & 78.4 & 77.5 & 65.8 & 57.2 & 80.2 & 72.7 & 60.7 & 83.3 & 71.8 \\
    TPDS~\cite{tang2024source}&  IJCV24 & $\checkmark$ &  59.3 & \textbf{80.3} & 82.1 & 70.6 & 79.4 & \textbf{80.9} & 69.8 & 56.8 & 82.1 & 74.5 & 61.2 & 85.3 & 73.5 \\
    Improved SFDA~\cite{Mitsuzumi2024Understanding} & CVPR24   & $\checkmark$  & \textbf{60.7}  & 78.9 & 82.0 & 69.9 & 79.5 & 79.7 & 67.1 & \textbf{58.8} & 82.3 & 74.2 & 61.3 & \textbf{86.4} & 73.4 \\
    \textbf{ARFNet}~(\textbf{Ours})& --  & 
    $\checkmark$ & 58.7	& 77.2	& 82.5 & 	\textbf{75.8} &	78.5 &	80.2 &	\textbf{74.7} &	56.4 &	83.5 &	\textbf{78.8} &	58.7 &	85.5 &
    \textbf{74.2}  \\
    \bottomrule
    \end{tabular}
  }
  \label{tab:office-home}%
  \vspace{-4mm}
\end{table*}%

\subsubsection{Results of Office-Home} 
According to the results on the Office-Home dataset shown in Tab.~\ref{tab:office-home},  exploiting the ResNet-50 directly to evaluate the target domain adaptation without achieving the source, it can obtain a score of 60.2\%.
In comparison, the performance of our method is 74.2\% in the source-free setting. 
Obviously, it can surpass the vast majority of current source-free SOTA approaches such as C-SFDA, TPDS, and Improved SFDA.
Besides, with the same setting, our approach can outperform the SOTA benchmark TPDS and C-SFDA by 0.7\%. Besides, our proposed approach achieves the best performance on the transfer tasks are three subdomains, including 
Cl$\rightarrow$Ar, Pr$\rightarrow$Ar, 
Rw$\rightarrow$Ar, 
which proves the necessity of improving the effect of attention fusion and contrast in domain adaptation.

\subsubsection{Results of VisDA-C} As shown in Tab.~\ref{tab:visda-c}, we mainly compare the results of SFDA on the ResNet-101 backbones.
First, we notice that the source-only model only achieves 48\% in the SFDA setting, which is 4.4\% inferior to the ResNet-101 backbone.
Therefore, for CNN-based methods in SFDA, a more rational design is needed. 
While the best result of source-free is 87.6\% by TPDS. Our performance can further improve the base to 88.9\%, which proves that our method is effective on large-scale domain adaptation datasets.
Besides, although our proposed approach only achieves the best performance on transfer tasks of car and truck, the average score still achieves the best, which demonstrates that the overall performance of our approach behaves well on large-scale datasets.

\subsubsection{Results of DomainNet-126}
To further validate the effectiveness of our approach on a large-scale dataset, we evaluate our approach with multiple SOTA algorithms in SFDA settings. 
If ResNet-50 is used as the backbone directly, its performance is only 54.7\%. However, in the Source-only setting, the adaptation performance can be improved only by 3.4\%.
The conventional methods like SHOT and NRC using pseudo-label evaluation can improve the performance to 68\%. Among these methods, the GKD approach showed state-of-the-art results, reaching a peak accuracy of 68.7\%.
Unlike traditional methods, our approach utilizes improved models with the proposed strategies in the data-free setting, achieving 70.3\% and outperforming other methods on three transfer tasks~(e.g., C$\rightarrow$P, C$\rightarrow$R, and C$\rightarrow$S, etc) to reach the latest SOTA.
\begin{table}[!htbp]
  \vspace{-5mm}
  \centering
  \caption{The ablation study of our approaches with different components. `+' denotes the add operation, DN-126 indicates DomainNet-126, while CUBs indicates CUB-Paddings.}
  \resizebox{\linewidth}{!}{
    \begin{tabular}{lcccccc}
    \toprule
    Method &  Office-31 & Office-Home & VisDA-C & DN-126 & CUBs & Avg. \\
    \midrule
    Source-Only & 79.3 & 60.2 &  48.0 & 58.1 & 48.7 & 58.9 \\
    \midrule
    $\mathcal{L}_{im}$~(Baseline) & 88.7 & 72.1   & 83.1 & 67.5 & 61.3 & 74.5 \\
    +  $\mathcal{L}_{ssd}$  & 89.6 & 73.5  & 86.3 & 68.8 & 62.7 & 76.8 \\
    +  $\mathcal{L}_{gac}$  & 89.1 & 72.9  &  85.7 & 69.7  & 64.8 & 76.4 \\
    +  $\mathcal{L}_{ssd}$ + $\mathcal{L}_{gac}$  & \textbf{90.8} & \textbf{74.2}   & \textbf{88.9} & \textbf{70.3} & \textbf{65.6} & \textbf{77.9} \\
    \bottomrule
    \end{tabular}%
    \label{tab:ablation_study}%
  }
\end{table}%
\subsection{Ablation Study}
\subsubsection{Effect of Components} To validate the effectiveness of the different components of our method, we perform ablation experiments on the five datasets for each component.
The corresponding results are reported in Tab.~\ref{tab:ablation_study}.
In the second row, we train our baseline by the information maximization loss $\mathcal{L}_{im}$ and obtain 88.7\%, 72.1\%, 83.1\%, 67.5\%, and 61.3\% on the five datasets, respectively.
If directly utilizing the self-supervised self-distillation approach $\mathcal{L}_{ssd}$ to supervise the training, the results can be improved by approximately 1.7\% on average. This demonstrates that dynamic centroid evaluation can further improve the confidence of categories. 
When the approach $\mathcal{L}_{gac}$ is utilized to compare the global and local attention, the results compared to the baseline can be improved by about 1.9\% on average, which demonstrates that the perceptual capabilities of different categories are improved effectively.
Finally, both $\mathcal{L}_{ssd}$ and $\mathcal{L}_{gac}$ are employed to verify our approach, which can improve the baseline by about 3.4\%.
This demonstrates that the proposed three components are critical for our approach to perform well on SFDA.
\begin{table}[!htbp]
  \vspace{-3mm}
  \centering
  \caption{Influence of Hyper-parameters $\alpha$ and $\beta$.}
    \begin{tabular}{c|cccccc}
      \toprule
      $\beta$ = 0, $\alpha$ = & 0.01  & 0.1   & 0.5  &  0.7 & 1.0  & 5.0 \\
      \hline
      Acc.~(\%)  & 73.2  & 73.3  & 73.5 & 74.0 & \textbf{74.2}   & 73.4  \\
      \midrule
      $\alpha$ = 1.0, $\beta$ = & 0.1   & 0.3 & 0.5  & 1.0   & 5.0   & 10 \\
      \hline
      Acc.~(\%)  & 74.4  & 74.9 & \textbf{75.8} & 75.4  & 75.1  & 75.2 \\
      \bottomrule
    \end{tabular}%
  \label{tab:affect_ab}%
\end{table}%

\subsubsection{Hyperparameters Analysis}
In our total loss function (Eq.~\ref{eq:final_loss}), $\alpha$ and $\beta$ are the major hyperparameters for balancing the loss terms of our framework. To analyze their effect on the performance, we conduct a series of ablation studies on the Cl$\rightarrow$Ar task of the Office-Home dataset. As shown in the Tab.~\ref{tab:affect_ab}, when $\beta$ is set to 0, adjusting the $\alpha$ from 0.01 to 5.0. This means the model only exploits the $\mathcal{L}_{ssd}$ to self-supervise the training. When $\alpha$ is set to 1.0, achieving the best result of 74.2\%. Based on the above, we set $\alpha=1.0$, adjusting the $\beta$ from 0.1 to 10, showing the results in the Tab.~\ref{tab:affect_ab}. The $\beta$ controls the feature contrast loss $\mathcal{L}_{gac}$ to discriminate the variations between intra-class and inter-class. When the $\beta$ is small, the performance of the model will be degraded, due to the inability to distinguish between the differences. However, when the $\beta$ is too large, the model overdistinguishes differences, which also affects performance. Consequently, a reasonable choice of $\beta$ is 0.5, which achieves the best score of 75.8\%.

\begin{figure}[!htbp]
  \centering
  \vspace{-2mm}
   \subfloat[$\tau$]{
      \includegraphics[width=0.45\linewidth]{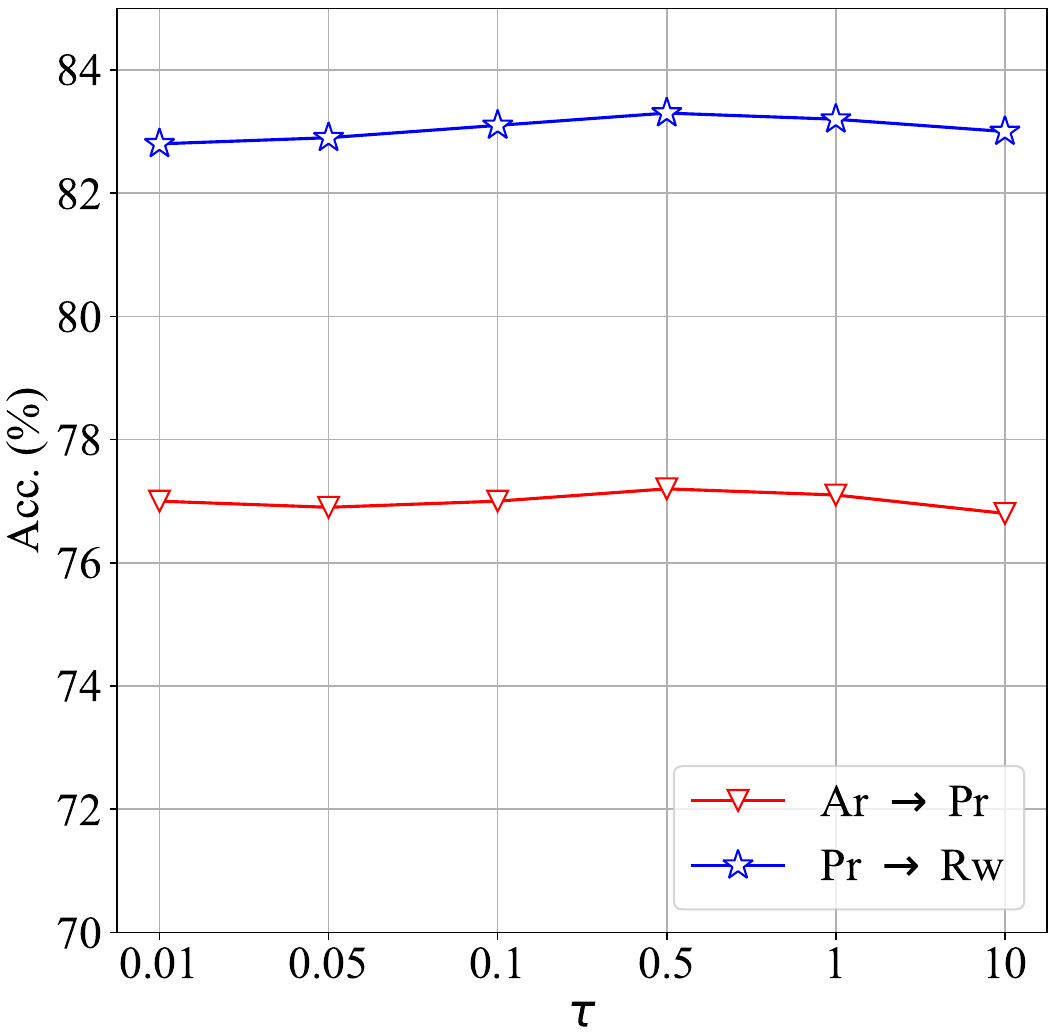}
      \label{fig:param_a}
   }
   \hfill
   \subfloat[$\lambda$]{
      \includegraphics[width=0.45\linewidth]{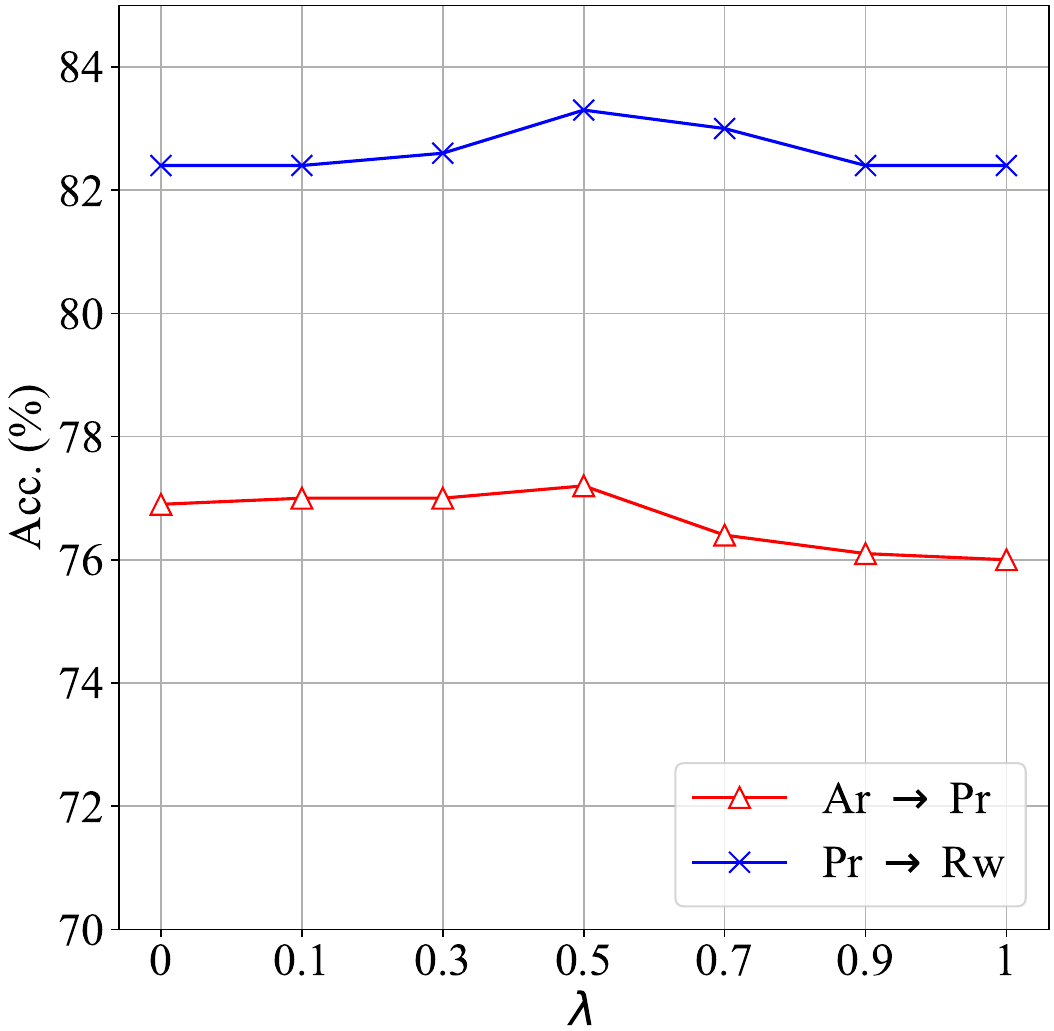}
      \label{fig:param_b}
   }
  \caption{Sensitivity of hyper-parameter  $\tau$ and $\lambda$ in specific tasks~(Ar$\rightarrow$Cl and Rw$\rightarrow$Cl) on Office-Home Dataset.}
  \label{fig:param_ab}
\end{figure}

%
\begin{table*}[!htbp]
  \centering
  \caption{Accuracy (\%) of different methods based on ResNet-101 unsupervised SFDA in the VisDA-C dataset.}
  \resizebox{\linewidth}{!}{
    \begin{tabular}{ll|c|ccccccccccccc}
    \toprule
    Method & Venue & S.F. & plane & bicycle & bus & car & horse & knife & motor & person & plant & skate & train & truck & Avg. \\
    \midrule
    Source-only & CVPR16 & $\checkmark$ & 64.1 & 24.9 & 53.0 & 66.5 & 67.9 & 9.1 & 84.5 & 21.1 & 62.8 & 29.8 & 83.5 & 9.3 & 48.0 \\
    3C-GAN~\cite{li2020model} & CVPR20 & $\checkmark$ & 94.8& 73.4& 68.8 &74.8& 93.1& 95.4 &88.6 & 84.7 &89.1& 84.7& 83.5 &48.1 & 81.6 \\    
    SHOT~\cite{liang2020we} & ICML20 & $\checkmark$ & 94.3 & 88.5& 80.1& 57.3 &93.1 &94.9& 80.7& 80.3& 91.5 &89.1 &86.3& 58.2  & 82.9\\     
    SFDA~\cite{kim2021domain} & AI21 & $\checkmark$ & 86.9 & 81.7 & 84.6 & 63.9 & 93.1& 91.4 &86.6 &71.9 &84.5 &58.2 &74.5& 42.7 & 76.7 \\    
    A$^2$Net~\cite{xia2021adaptive} & ICCV21 & $\checkmark$ & 94.0 & 87.8& 85.6& 66.8& 93.7& 95.1& 85.8& 81.2 &91.6& 88.2& 86.5& 56.0 & 84.3\\    
    G-SFDA~\cite{yang2021generalized} & ICCV21 & $\checkmark$ & 96.1 & 88.3 &85.5 &74.1 &97.1 & 95.4 & 89.5 & 79.4& 95.4& 92.9& 89.1& 42.6& 85.4 \\
    SFDA-DE~\cite{ding2022source} & CVPR22 & $\checkmark$ & 95.3 & 91.2 & 77.5 & 72.1 & 95.7 & \textbf{97.8}  & 85.5 & 86.1 & 95.5 & 93.0 & 86.3 & 61.6 & 86.5 \\   
    DaC~\cite{zhang2022divide}  & NIPS22 & $\checkmark$ & 96.6 & 86.8 & 86.4 & 78.4 & 96.4 & 96.2 & \textbf{93.6} & 83.8 &  \textbf{96.8} & 95.1 & 89.6 & 50.0 & 87.3 \\   
    PLUE~\cite{Litrico_2023} & CVPR23  & $\checkmark$ &  94.4 & \textbf{91.7} & 89.0 & 70.5 & 96.6 & 94.9 & 92.2 &  \textbf{88.8} & 92.9 &  \textbf{95.3} & 91.4 & 61.6 & 88.3\\
    C-SFDA ~\cite{Karim2023sfda} & CVPR23  & $\checkmark$  & \textbf{97.6} & {88.8} & {86.1} & {72.2} & {97.2} & 94.4 & 92.1 & {84.7} & {93.0} & {90.7} &  \textbf{93.1} & \textbf{63.5} & 87.8 \\  
    ASOGE~\cite{cui2024adversarial} & TCSVT24 & $\checkmark$ & 94.9 & 84.3 & 76.8 & 54.3 & 94.9 & 93.4 & 86.0 & 85.0 & 87.2 & 90.0 & 86.7 & 62.7 & 83.2 \\
    TPDS~\cite{tang2024source} & IJCV24 & $\checkmark$ &  97.6 & 91.5 & \textbf{89.7} & 83.4 & \textbf{97.5} & 96.3 & 92.2 & 82.4 & 96.0 & 94.1 & 90.9 & 40.4 & 87.6 \\
    Improved SFDA~\cite{Mitsuzumi2024Understanding} & CVPR24   & $\checkmark$  & 97.5 & 91.4 & 87.9 & 79.4 & 97.2 & 97.2 & 92.2 & 83.0 & 96.4 & 94.2 & 91.1 & 53.0 & 88.4 \\
    \textbf{ARFNet~(Ours)}  & -- & $\checkmark$ &  97.4 &	91.6 &	89.1 &	\textbf{84.1} &	95.9 &	93.9 &	87.7 &	86.8 &	95.4 &	94.1 &	89.1 &	62.2 &   \textbf{88.9} \\
    \bottomrule
    \end{tabular}%
    }
  \label{tab:visda-c}%
  \vspace{-4mm}
\end{table*}%

\subsubsection{Sensitivity Analysis} 
Besides, $\tau$ and $\lambda$ are the inner parameters of our methods. To verify its robustness, we adjust two parameters $\tau$ $\in$ $\left \{ 0.01, 0.05, 0.1, 0.5, 1, 10 \right \}$ and $\lambda \in \left \{ 0, 0.1, 0.3, 0.5, 0.7, 0.9, 1 \right \}$, showing the sensitivity of the model.
As depicted in Fig.~\ref{fig:param_a}, from which we observe that $\tau$ of $\mathcal{L}_{gac}$ is robust for our approach, and 0.05 to 1.0 is reasonable. The $\lambda$ balance factor of dynamic centroids indicates the selection of dynamic centroids and plays a key role in determining the center point. When the $\lambda$ is given 0, it implies using a traditional static approach for centroids assessment.
For the tasks of Ar$\rightarrow$Pr and Pr$\rightarrow$Rw, the best performance is achieved when the $\lambda$ is set to 0.5, as shown in Fig.~\ref{fig:param_b}.

\subsection{Visualization}
To explore the alignment effect of the final source domain and target domain, we demonstrate the principle and effect of our experiment through visual analysis. It contains two main aspects, the visualization of the attention map by grad-cam~\footnote{https://github.com/jacobgil/pytorch-grad-cam} and the visualization of feature alignment by t-SNE.

\subsubsection{t-SNE}
\begin{figure}
  \centering
  \vspace{-2mm}
   \subfloat[Before Adaptation]{
    \includegraphics[width=0.43\linewidth]{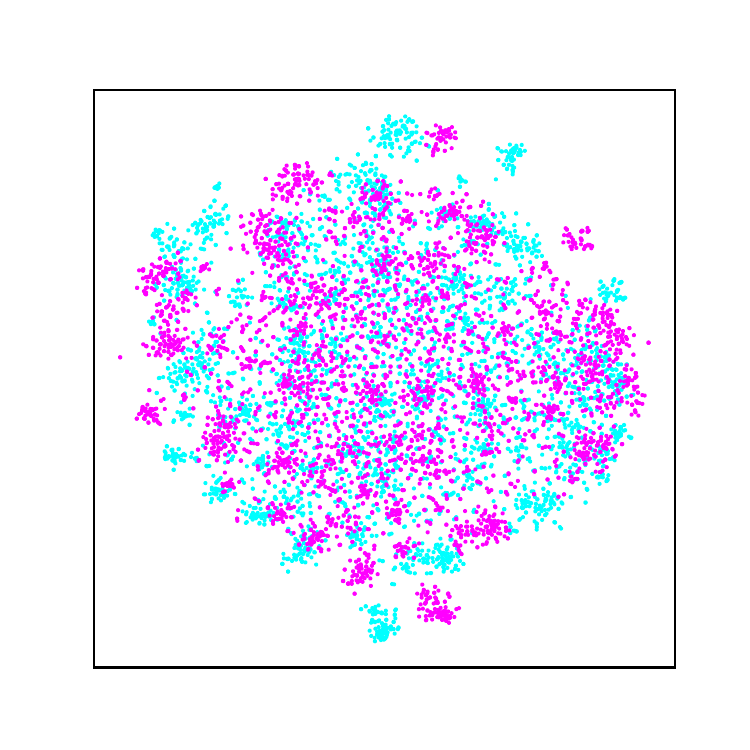}
    \label{fig:tsne_a}
   }
   \subfloat[After Adaptation]{
    \includegraphics[width=0.43\linewidth]{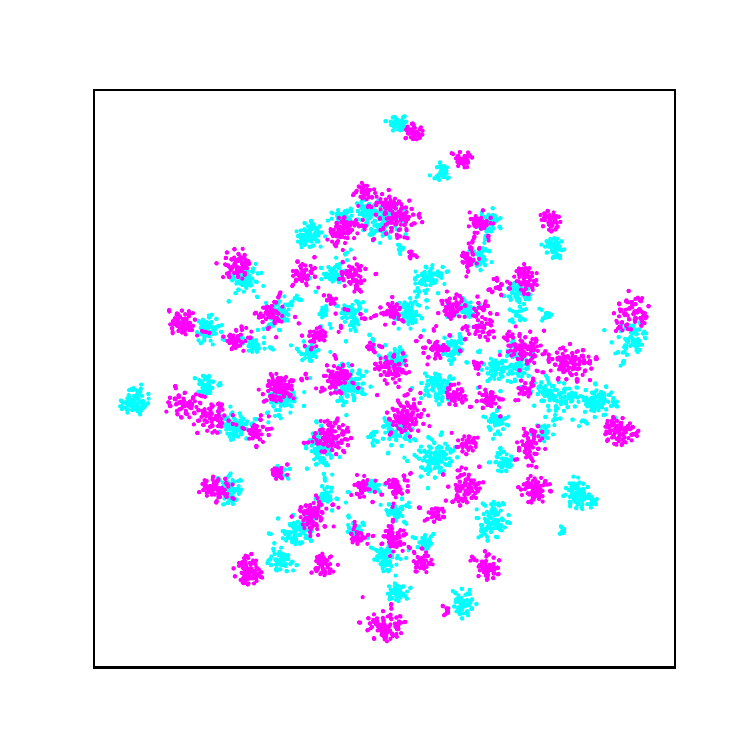}
    \label{fig:tsne_b}
   }
  \caption{t-SNE visualization for domain adaptation on Office-Home (Ar $\rightarrow$ Cl), Red denotes the source domains, while black denotes the target domains.}
  \label{fig:tsne_ab}
  \vspace{-4mm}
\end{figure}

To demonstrate the effect of different methods on domain alignment, we utilize t-SNE to visualize the distributions of feature representations, which are obtained from the penultimate layer in both source and target domains.
As can be seen from Fig.~\ref{fig:tsne_a}, the sample features of the same category are more dispersed before adaptation.
This might be due to the severe domain shift problem with the source data.
After adaptation, we can observe that the category distance is significantly reduced and the sample categories are distributed more clearly.

\begin{table*}[!htbp]
  \caption{ Accuracy (\%) of different methods based on ResNet-50 unsupervised SFDA on {\bf DomainNet-126}.  }
  \resizebox{\linewidth}{!}{
  \centering
  \begin{tabular}{ l l|c |c c c  c c c c c c c c c c }
    \toprule
      Method &{Venue} &{\bf S.F.}
      &C$\to$P &C$\to$R &C$\to$S
      &P$\to$C &P$\to$R &P$\to$S    
      &R$\to$C &R$\to$P &R$\to$S  
      &S$\to$C &S$\to$P &S$\to$R &Avg.\\
      \midrule
      Source-only~\cite{he2016deep}   & CVPR16    &\checkmark   & 48.8	& 62.7	& 50.5 &	59.4 &	64.6 &	50.5 &	57.6 &	76.0 &	50.4	&60.1 &	54.2 &	62.4 &	58.1 \\
      SHOT~\cite{liang2020we}     &ICML20  &\checkmark &63.5 &78.2 &59.5 &67.9 &81.3 &61.7 &67.7 &67.6 &57.8 &70.2 &64.0 &78.0 &68.1 \\
      GKD~\cite{tang2021model}    &IROS21  &\checkmark  &61.4 &77.4 &60.3 &\textbf{69.6} &\textbf{81.4} &\textbf{63.2} &68.3 &68.4 &59.5 &71.5 &65.2 &77.6 &68.7 \\ 
      NRC~\cite{yang2021nrc}      &NIPS21  &\checkmark  &62.6 &77.1 &58.3 &62.9 &81.3 &60.7 &64.7 &69.4 &58.7 &69.4 &65.8 &78.7 &67.5 \\
      AdaCon~\cite{chen2022contrastive}    &CVPR22  &\checkmark  &60.8 &74.8 &55.9 &62.2 &78.3 &58.2 &63.1 &68.1 &55.6 &67.1 &\textbf{66.0} &75.4 &65.4 \\
      CoWA~\cite{lee2022confidence}        &ICML22  &\checkmark &64.6 &80.6 &60.6 &66.2 &79.8 &60.8 &\textbf{69.0} &67.2 &60.0 &69.0 &65.8 &79.9 &68.6\\
      PLUE~\cite{Litrico_2023}        &CVPR23  &\checkmark &59.8 &74.0 &56.0 &61.6 &78.5 &57.9 &61.6 &65.9 &53.8 &67.5 &64.3 &76.0 &64.7  \\
      TPDS~\cite{tang2024source}           &IJCV24  &\checkmark  &62.9 &77.1 &59.8 &65.6 &79.0 &61.5 &66.4 &67.0 &58.2 &68.6 &64.3 &75.3 &67.1 \\
      \textbf{ARFNet~(Ours)}          & -- &\checkmark  & \textbf{66.9}  &	 \textbf{80.7}&	\textbf{62.9}& 67.0&	71.7&	60.5&	67.0&	\textbf{83.1}&	\textbf{64.4}&	\textbf{73.2}&	65.8 & \textbf{80.7}& \textbf{70.3} \\
    \bottomrule
  \end{tabular}
  }
  \label{tab:dn}
  \vspace{-3mm}
\end{table*}

\subsubsection{Attention Map}
\begin{figure}[!htbp]
    \centering
    \includegraphics[width=0.9\linewidth]{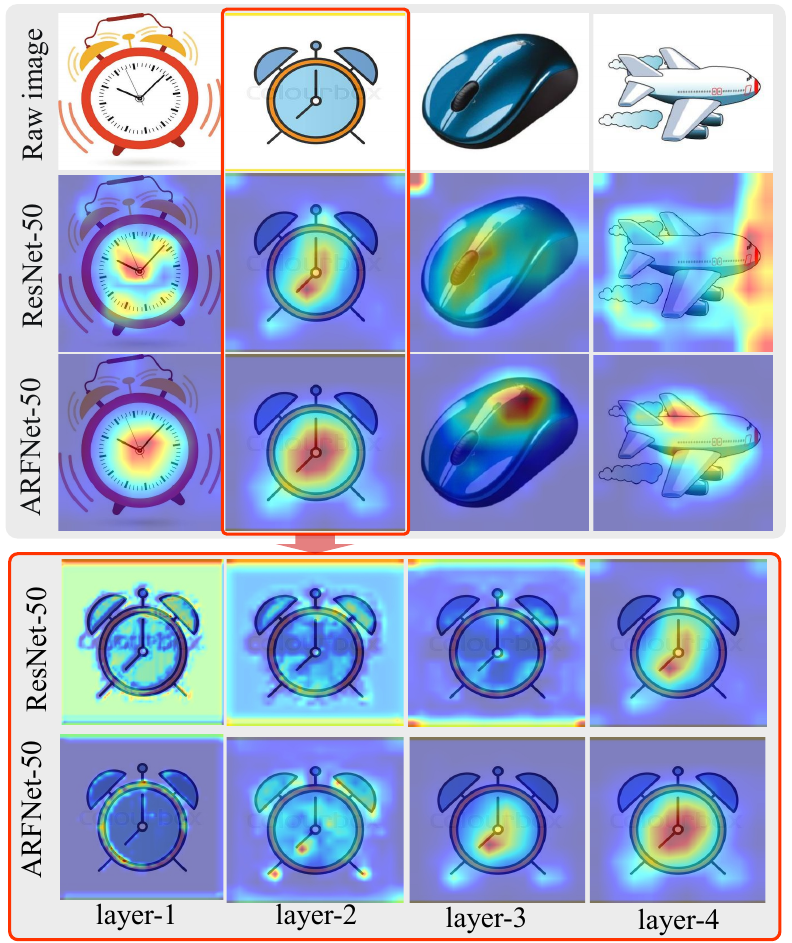}
    \vspace{-2mm}
    \caption{Attention maps of images about clock, mouse, and aircraft in the DomainNet-126 dataset. }
    \label{fig:attention}
    \vspace{-3mm}
\end{figure}

The attention maps depict the intensity of focus on the target region, where warmer colors indicate a higher level of attention directed toward that area.
According to attention maps shown in Fig.~\ref{fig:attention}, the original `Source Only' can hardly pay attention to the main regions, due to the original ResNet model having a weak ability to focus on the target object.
Although some previous research has focused on the objects, it is not comprehensive enough.
Compared with the former methods, our proposed method ARFNet can accurately capture discriminative region features, due to the residual fusion.
For example, the ResNet-50 only focuses on very few areas of the alarm clock, but the ARFNet-50 can focus on a much larger area than the ResNet-50.

\subsection{Extended Experiments}
\subsubsection{Model Scale Expansion Experiments}
In this paper, we primarily employ the ResNet family of networks as the backbone for our experiments, which focus on models with smaller parameters. Recently, large models utilizing the ViT network backbone have demonstrated remarkable success across various domains. To further validate that our proposed method is also effective for larger parameter models, we conduct additional experiments using ViT-based architectures. In each layer of the Multi-head Self-Attention (MHSA) layers, we first derive attention features from the $l$-th layer, denoted as $f^{l}_{t} \in \mathcal{R}^{S \times D}$, where $S$ represents the sequence length of patch tokens and $D$ signifies the dimension of the latent embedding. Subsequently, we reshape the image size into three dimensions as $\mathcal{R}^{D \times H \times W }$. Next, we inject the AEM blocks into the ViT backbone and input the reshaped features into the blocks to obtain channel-wise and spatial-wise attention features. 
Finally, we execute the entire workflow with proposed strategies and compare the results with current SOTA baselines in both UDA and SFDA settings on two representative datasets: Office-Home and VisDA-C, as illustrated in Tab.~\ref{tab:office-home_visda}.

\begin{table}[!htbp]   
  \color{black}
  \centering
  \caption{Accuracy (\%) of different methods of unsupervised SFDA on the Cub-Paintings dataset.}
    \label{tab:cub-paintings}%
      \begin{tabular}{llcccc}
      \toprule
        Method & Venue & S.F. & C$\rightarrow$P   & P$\rightarrow$C   & Avg.   \\
      \midrule         
        Source-only & CVPR16 & $\checkmark$ & 55.4  & 42.0  & 48.7 \\

        SHOT~\cite{liang2020we}     &ICML20  & $\checkmark$      & 65.8  & 59.6  & 62.7 \\

        GKD~\cite{tang2021model}      &IROS21  & $\checkmark$   & 65.4 &	57.8	& 61.6 \\   %

        NRC~\cite{yang2021nrc}      &NIPS21  &\checkmark   & 65.3  & 59.4  & 62.4 \\  %

        PLUE ~\cite{Litrico_2023}        &CVPR23  &\checkmark  & 54.9  & 48.9  & 51.9\\ 

        TPDS ~\cite{tang2024source}        &IJCV24  &\checkmark  & 68.3  & 60.0 & 64.2 \\

        \textbf{ARFNet}   & --  &\checkmark  & \textbf{69.6}  & \textbf{61.3}  & \textbf{65.6} \\

        \midrule
        DIFO-RN~\cite{tang2024sfda} &  CVPR24 & $\checkmark$      & 63.6  & 54.8  & 59.2 \\

        ProDe-RN~\cite{tang2025proxy} &  ICLR25   & $\checkmark$   & 68.7  &  61.8  &  65.3  \\

        DIFO-B32~\cite{tang2024sfda} &  CVPR24 & $\checkmark$      & 65.4  & 58.0  & 61.7 \\

        ProDe-B32~\cite{tang2025proxy} &  ICLR25   & $\checkmark$   & 69.4  & 63.4  & 66.4 \\

        \textbf{ARFNet-RN} & -- & $\checkmark$  & 69.8  & 61.6 & 65.7 \\

        \textbf{ARFNet-B32} & -- & $\checkmark$   & \textbf{70.2}  & \textbf{64.0} & \textbf{67.1} \\
      \bottomrule
      \end{tabular}%
    \vspace{-4mm}
\end{table}%

In the Office-Home dataset, employing the ViT-B/16 network as the backbone within a domain adaptation (UDA) setting can achieve a score of 73.6\%. Recent research~\cite{xu2021cdtrans,sun2022safe,yang2023tvt} has explored new domain adaptation paradigms using the ViT backbone, as evidenced by studies such as CDTrans, SSRT, and TVT. These approaches have successfully increased the baseline score to 85.4\% by utilizing the SSRT method. Additionally, in the VisDA dataset, SSRT enhances the baseline performance of the ViT by 22.7\%.
In the SFDA setting, research exploiting ViT as a backbone is relatively scarce. Through experiments, if directly using the ViT-B/16 model for SFDA, we can obtain a score of 73.2\%. The TPDS based on the ViT-B/16 model achieved the SOTA performance, obtaining 81.8\% and 87.7\% on the Office-Home and VisDA-C datasets, respectively.
Our approaches can also extend to the ViT-based model and surpass other S.F. methods, achieving SOTA performance of 82.6\% and 88.5\% on both datasets.

Recent research trends highlight the integration of CLIP frameworks with ResNet (*-RN) and ViT (*-B32) backbones to create a domain-specific prompt learning paradigm. This approach accomplishes vision-language domain adaptation by aligning semantics across modalities. 
It effectively addresses the semantic ambiguity issues found in traditional adaptation methods and has demonstrated significant performance improvements, with baseline enhancements reaching up to 74.7\% and 82.9\%, respectively.
In the UDA setting, current approaches exploit CLIP with ResNet-50/101 as backbones for Office-home and VisDA-C, respectively, as shown in the lower part of the Tab.~\ref{tab:office-home_visda}. 
PADCLIP-RN and DAMP-RN validate the effectiveness of vision-language foundation models in UDA and address the challenges of cross-domain adaptation in different directions, achieving SOTA performance scores of 78.2 and 88.5 on the Office-home and VisDA-C datasets, respectively.
In the SFDA setting, current research DIFO and ProDe exploit CLIP with ResNet and ViT-B/32 models and report SOTA performances on both datasets.
By applying our method to existing frameworks to verify their performance. Our method can further enhance the current outcomes on both ResNet and ViT-based models.

\begin{table*}[!htbp]
  \color{black}
  \centering
  \vspace{-3mm}
  \caption{Accuracy (\%) of different methods of unsupervised SFDA on the Office-Home and VisDA-C dataset.}
  \resizebox{\linewidth}{!}{
    \begin{tabular}{llcccccccccccccccc}
      \toprule    
      \multirow{2}{30pt}{\centering Method} & \multirow{2}{*}{\centering Venue} & \multirow{2}{*}{\centering S.F.} && \multicolumn{12}{c}{\textbf{Office-home}} && \textbf{VisDA-C} \\
      \cmidrule{4-16} \cmidrule{18-18}
      &&& Ar$\rightarrow$Cl & Ar$\rightarrow$Pr &  Ar$\rightarrow$Rw  &  Cl$\rightarrow$Ar &  Cl$\rightarrow$Pr  &  Cl$\rightarrow$Rw & Pr$\rightarrow$Ar  & Pr$\rightarrow$Cl  &  Pr$\rightarrow$Rw  & Rw$\rightarrow$Ar  & Rw$\rightarrow$Cl  &  Rw$\rightarrow$Pr & Avg. && Sy $\rightarrow$ Re \\
      
      \midrule
      ViT-B16~\cite{dosovitskiy2020image} & ICLR21 & $\times$ & 51.4 &	81.1 &	85.4 &	73.6 &	82.2 & 83.0 & 73.6 & 50.8 & 87.2 & 78.2 & 50.1 & 86.4 & 73.6 & & 66.1 \\
      CDTrans~\cite{xu2021cdtrans} & ICLR22 & $\times$ & 68.8 & 85.0 & 86.9 &81.5 & 87.1 & 87.3 & 79.6 & 63.3 & 88.2 & 82.0 & 66.0 & 90.6 & 80.5 & & 88.4 \\
      SSRT~\cite{sun2022safe} & CVPR22 & $\times$ &  75.2 & 89.0 & 91.1 & 85.1 & 88.3 & 90.0 & 85.0 & 74.2 & 91.3 & 85.7 & 78.6 & 91.8 & 85.4 & & 88.8 \\
      TVT~\cite{yang2023tvt} & WCACV23 & $\times$ & 74.9 & 86.8 & 89.5 & 82.8 & 88.0 & 88.3 & 79.8 & 71.9 & 90.1 & 85.5 & 74.6 & 90.6 & 83.6 & & 86.7 \\
      \midrule
      Source-only~\cite{dosovitskiy2020image} & ICLR21 & $\checkmark$ & 66.2 &	84.3 &	86.6 &	77.9 & 	83.3 &	84.3 &	76.0 & 	62.7 & 	88.7  &	 80.1  &	66.2  & 	88.6  &	78.8 &  & 73.2\\
      TransDA~\cite{yang2023self}& APIN23 & $\checkmark$ & 67.5  & 83.3  & 85.9  & 74.0  & 83.8  & 84.4  & 77.0  &  \textbf{68.0}  & 87.0  & 80.5  & 69.9  & 90.0  & 79.3 & & 83.0 \\
      DSiT (B16)~\cite{sanyal2023domain} & ICCV23  & $\checkmark$ & \textbf{69.2} & 83.5 & 87.3 &  80.7 & 86.1 & 86.2 & 77.9 & 67.9 & 86.6 & 82.4 & 68.3 & 89.8 & 80.5 & & 87.6 \\
      TPDS (B16)~\cite{tang2024source} & IJCV24  & $\checkmark$ &  65.1	&  \textbf{90.3} &	 \textbf{90.5} &	81.2 &	 89.4 &	\textbf{89.6}	& 80.8 &	61.0 &	90.6 &	82.3 &	61.4 &	90.7 &	81.1 &  & 87.8 \\
      \textbf{ARFNet (B16)} & -- & $\checkmark$  & 
      64.3 &	{89.5} &	{89.9} &	\textbf{82.2} &	\textbf{91.0} &	89.4 &	\textbf{82.0} &	67.7 &	\textbf{90.2} &	\textbf{82.5} &	\textbf{70.8} &	\textbf{91.2} &	\textbf{82.6} & & \textbf{88.5} \\ 
      \midrule
      CLIP-B32~\cite{radford2021learning} & ICML21 & $\times$ & 54.7 &	84.2 &	82.5 &	73.8 &	80.5	& 83.2 &	74.7 &	56.4 &	83.3	& 55.8	& 81.3 &	85.5 &	74.7  &  & 82.9 \\
      DAPrompt-RN~\cite{ge2025domain} & TNNLS25 & $\times$ & 54.1 & 84.3 & 84.8 & 74.4 & 83.7 & 85.0 & 74.5 & 54.6 & 84.8 & 75.2 & 54.7 & 83.8 & 74.5 & & 86.9 \\
      PADCLIP-RN~\cite{lai2023pseudo} & ICCV23 & $\times$ & 57.5 & 84.0 & 83.8 & 77.8 & 85.5 & 84.7 & 76.3 & 59.2 & 85.4 & 78.1 & 60.2 & 86.7 & 76.6 & & 88.5 \\
      ADCLIP-RN~\cite{singha2023ad} & ICCVW23 & $\times$ & 55.4 & 85.2 &85.6 & 76.1 & 85.8 & 86.2 & 76.7 & 56.1 & 85.4 & 76.8 & 56.1 & 85.5 & 75.9 & & 87.7 \\
      PDA-RN~\cite{bai2024prompt} & AAAI24 &$\times$& 55.4 & 85.1& 85.8& 75.2& 85.2& 85.2& 74.2& 55.2& 85.8& 74.7& 55.8 &86.3 &75.3 & & 86.4 \\
      DAMP-RN~\cite{du2024domain} & CVPR24 &$\times$ & 59.7 & 88.5 & 86.8 & 76.6 & 88.9  & 87.0 & 76.3 & 59.6 & 87.1 & 77.0 & 61.0 & 89.9 &78.2 &  & 88.4 \\
      \midrule
      DIFO-RN~\cite{tang2024sfda} & CVPR24 & $\checkmark$ & 62.6 & 87.5 & 87.1 & 79.5 & 87.9 & 87.4 & 78.3 & 63.4 & 88.1 &  80.0 & 63.3 & 87.7 & 79.4 & & 88.6 \\

      DIFO-B32~\cite{tang2024sfda} & CVPR24 & $\checkmark$ & 70.6 & 90.6 & 88.8 & 82.5 & 90.6 & 88.8 & 80.9 & 70.1 & 88.9 & \textbf{83.4} & 70.5 & 91.2 & 83.1 &  & 90.3 \\

      ProDe-RN~\cite{tang2025proxy} & ICLR25 & $\checkmark$ & 64.0 & 90.0 & 88.3 &  81.1 & 90.1 & 88.6 & 79.8 & 65.4 & 89.0 & 80.9 & 65.5 & 90.2 & 81.1 &  & 88.7 \\

      ProDe-B32~\cite{tang2025proxy} & ICLR25 & $\checkmark$ & 72.7 & 92.3 & 90.5 &  82.5 &  91.5 & \textbf{90.7} & 82.5 &  72.5 &  90.8 &  83.0 &  72.6 &  92.2 & 84.5 &  & 91.0 \\  

      \textbf{ARFNet-RN} & -- & $\checkmark$  & 
      65.6 &	91.6 &	91.2 &	82.2 &	91.7 &	90.2 &	82.9 &	62.3 &	90.7 &	82.5 &	72.9	& 91.5 &	83.0 &  & 89.5 \\  
      \textbf{ARFNet-B32} & -- & $\checkmark$  &  \textbf{73.3} &	\textbf{92.9} &	\textbf{91.4}	& \textbf{83.2}	& \textbf{91.8}	& 90.5	& \textbf{84.6}	& \textbf{72.9}	& \textbf{91.2} &	82.8	& \textbf{74.8} & \textbf{92.5} &\textbf{85.2}  &  &\textbf{91.6} \\  
      \bottomrule
    \end{tabular}
  \label{tab:office-home_visda}
  }
  \vspace{-2mm}
\end{table*}%

\subsubsection{Effectiveness on Fine-grained Cub-Paintings Dataset}
  To further validate the effectiveness of our method, we conduct experiments on the Cub-Paintings dataset~\cite{PAN}, which is a fine-grained dataset with 200 categories of web and paintings. 
  The results are shown in Tab.~\ref{tab:cub-paintings}.
  In the top section of the table, we utilize ResNet-50 as the backbone and compare our method with other S.F. methods. The results show that our method achieves a classification accuracy of 69.6\% on the adaptation of C$\rightarrow$P  and 61.3\% on the adaptation of P$\rightarrow$C, with an average accuracy of 65.7\%. 
  In the bottom section, we utilize ReNet-50/CLIP and ViT-B32/CLIP as the backbones and compare our method with other S.F. methods. The results show that our method achieves a classification accuracy of 70.2\% on the adaptation of C$\rightarrow$P and 64.0\% on the adaptation of P$\rightarrow$C, with an average accuracy of 67.1\%.
  Compared to the previous S.F. methods, our method achieves the best performance on both adaptation tasks.
  This demonstrates that our method effectively captures the discriminative areas in fine-grained images and reduces negative transfer. Hence, it can adapt to fine-grained datasets and achieve competitive performance compared to other state-of-the-art methods.

\begin{figure}[!tpb] 
  \centering 
    \includegraphics[width=\linewidth]{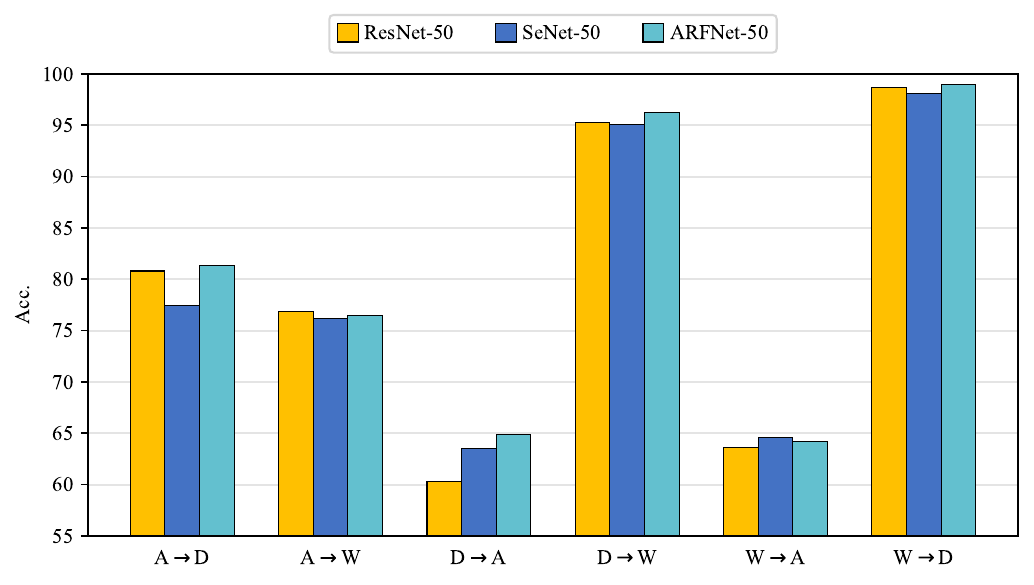}
    \caption{Effect of domain adaptation on Office-31 in `Source-only' scenarios. }
    \label{fig:target_da}  
  \vspace{-4mm}
\end{figure}

\subsection{Effectiveness of the ARFNet backbone}
We first train our proposed framework in the source domain in a supervised learning manner by the loss of cross entropy as in Eq.~\ref{eq:source_ce}, since the data and labels are accessible on the source domain.
For the design of our model inspired by ResNet~\cite{he2016deep} and SENet~\cite{hu2018squeeze}, we compare the proposed method with these two general backbones. 
As shown in the following Tab.~\ref{tab:model_effect_overhead} on the source domain of all five datasets, our approach can effectively improve the effectiveness of the model.
On the Office-31, Office-Home, DomainNet-126, and CUB-Paddings datasets, our proposed model ARFNet-50 can improve the baseline of ResNet-50 by 1.5\% on average, while on the VisDA-C, our model ARFNet-101 can improve the results of the model by 0.9\% on average. Similarly, for SENet-50, our ARFNet-50 can also improve both baselines by 1.1\% on average.

To further validate the effect of the model in the `Source-only' scenarios, we detailed each adaptation effect of the model on the Office-31 dataset. As exhibited in Fig.~\ref{fig:target_da}, our method achieves advantages in domain adaptation of both A$\rightarrow$ D, D $\rightarrow$ A, D $\rightarrow$ W, and W $\rightarrow$ D. 
Two group experiments demonstrate that our approach is capable of capturing essential discriminative features to improve supervised learning and effectively executing domain adaptation in `Source-only' scenarios. This plays a crucial role in mitigating the effects of model negative transfer during the process of adapting to the target domain.

Due to the introduction of the attention decomposition module in our method, we further evaluate and compare the different model parameters~(\textbf{Param.}) and computational overhead in Tab.~\ref{tab:model_effect_overhead}, i.e., the number of parameters and floating point operations per second~(\textbf{FLOPs}).
As shown in the table,  our proposed ARFNet demonstrates clear advantages over both ResNet and SENet in balancing computational efficiency and model performance. 
Compared to ResNet-50, ARFNet-50 achieves an obvious 1.1\% improvement in accuracy while only increasing the number of parameters by 3.8\% (26.52M vs. 25.56M) and computational overhead by 21.3\% (5.01G vs. 4.13G FLOPs). Similarly, ARFNet-101 improves accuracy by 0.9\% over ResNet-101 with a 2.2\% increase in parameters (45.51M vs. 44.55M) and a 10.9\% increase in FLOPs (8.74G vs. 7.87G).
Compared to SENet-50, ARFNet-50 achieves a 0.3\% improvement in accuracy while reducing parameters by 5.5\% (26.52M vs. 28.07M) and maintaining comparable FLOPs (5.01G vs. 4.14G). ARFNet-101 also achieves a 0.9\% improvement over SENet-101 with a 7.7\% reduction in parameters (45.51M vs. 49.29M) and only slightly higher FLOPs (8.74G vs. 7.88G).
This demonstrates that ARFNet effectively enhances model performance without excessively increasing computational costs, making it suitable for resource-constrained environments.

\begin{table*}[!ht]
  \color{black}
  \centering
  \caption{Compare the performance and overhead of ARFNet with traditional backbones.}
  \resizebox{\linewidth}{!}{
    \begin{tabular}{c|ccccccc|c|ccc}
      \toprule
        Model &  Office-31 & Office-Home & DomainNet-126 & CUB-Paddings & Avg. & FLOPs &  Param.  & Model & VisDA-C & FLOPs &  Param. \\ 
      \midrule
        ResNet-50 & 79.3 & 60.2  & 58.1 &  48.7 & 61.6 & $\sim$4.13G  & $\sim$25.56M & ResNet-101 & 48.0 & $\sim$7.87G & $\sim$44.55M \\ %
        
        SENet-50 & 78.9  &  60.8  &  57.4 & 50.7 & 62.0 & $\sim$4.14G  & $\sim$28.07M  & SENet-101 &  47.8  &  $\sim$7.88G &  $\sim$49.29M   \\ %
        
        \textbf{ARFNet-50}  & \textbf{80.4}  &   \textbf{61.1}   &  \textbf{58.6} & \textbf{52.1} &\textbf{63.1} &$\sim$5.01G  & $\sim$26.52M & \textbf{ARFNet-101}   &   \textbf{48.9} & $\sim$8.74G &  $\sim$45.51M  \\
      \bottomrule
    \end{tabular}%
    }
  \label{tab:model_effect_overhead}%
  \vspace{-4mm}
\end{table*}%
\subsection{Discussion, Deficiencies and Feature work}
\subsubsection{Discussion} Current research on UDA is very extensive. SFDA is a very challenging area in the UDA field due to the inaccessibility of source domain data. To solve the dilemma of source domains, most of the traditional approaches consider invariant features or proxy domains by feature-level design, and these approaches have reached a certain bottleneck. 
Thanks to current new architectures (e.g., Transformer), relevant research can further improve the performance of models, thus prompting us to revisit the main challenges of domain adaptation learning, namely noise disturbance and negative transfer, from a model design perspective. Although relevant experiments have demonstrated the effectiveness of the methodology, there are still some shortcomings.
\subsubsection{Deficiencies} While our explorations make some progress in mitigating negative transfer and noise reduction through MARF, GAC, and DCE, some limitations in application still remain. Primarily, current implementations have focused on typical classification tasks within established benchmarks, with broader research yet to be explored. Secondly, current research trajectories reveal significant directions of multi-modal UDA, which aims to capture modality-invariant features. Although this approach can significantly improve the effectiveness of model learning, the multi-model architecture further increases the computational complexity. Finally, current research primarily focuses on SFDA in a single-source setting, further exploration in multi-source settings is needed. 
\subsubsection{Future work}
Based on the above deficiencies, future research will focus on the following three directions. Building upon the current framework, future efforts will extend the methodology to cross-domain generalization in detection/segmentation tasks while integrating vision-language pretraining for multimodal adaptation. 
In UDA or SFDA, we will explore lighter-weight and more efficient model designs for complex large-scale model training, allowing models to be adapted for more applied practice.
Lastly, we will investigate more complex source-free adaptation methods in multi-source domain scenarios, enabling the model to learn better and integrate knowledge from various source domains effectively.

\subsection{Conclusion}
In this paper, we propose a novel framework ARFNet based on contrast learning for SFDA, which aims to alleviate negative transfer and domain shift during the process of adaptation. 
The main innovations of our approach are attributed to three technical components: MARF, GAC, and DCE.
For MARF, we improve the architecture of the general backbone by introducing the attention mechanism to learn the discriminative features and fuse them layer by layer progressively.
For GAC, which can further improve the
perceptual capabilities of different categories and distinguish discrepancies between intra-class and inter-class categories.
For DCE, which can accurately evaluate the
center of the feature and effectively mitigate domain drift for the target domain. 
Comprehensive experiments have demonstrated that our proposed approach is not only effective on traditional CNN architectures but can also be adapted to other new architectures.




 
\bibliographystyle{IEEEtran}
\bibliography{IEEEabrv,ref}
%

{
\vspace{-4cm}
\begin{IEEEbiography}[
    {\includegraphics[width=1.2in,height=1.25in,clip,keepaspectratio]{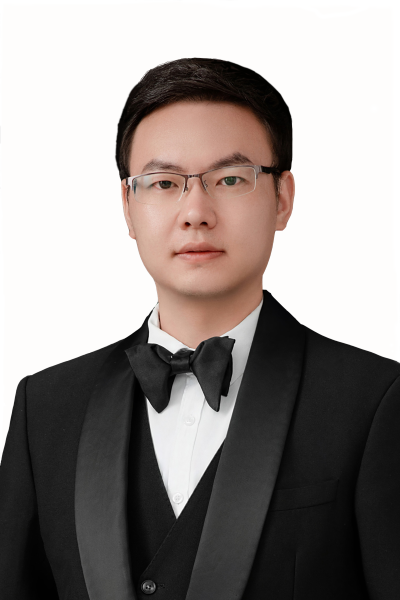}}]{Renrong Shao}
    received his Ph.D. degree in the School of Computer Science and Technology at East China Normal University, in 2024. He is currently a Lecturer and Postdoc Fellow in the Faculty of Military Health Services, Naval Medical University (Second Military Medical University), Shanghai. His research interests include computer vision, model compression, transfer learning and intelligent
healthcare.   
\end{IEEEbiography} 
}

{ 
\vspace{-4cm}
\begin{IEEEbiography}[   
    {\includegraphics[width=0.9in,height=1.25in,clip,keepaspectratio]{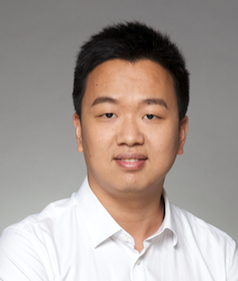}}]{Wei Zhang} 
    received his Ph.D. degree in computer science and technology from Tsinghua University, Beijing, China, in 2016. He is currently a professor in the School of Computer Science and Technology, East China Normal University, Shanghai, China. His research interests mainly include user data mining and machine learning applications. He is a senior member of China Computer Federation.
\end{IEEEbiography}
}

{
\vspace{-4cm}
\begin{IEEEbiography}[
    {\includegraphics[width=1in,height=1.25in,clip,keepaspectratio]{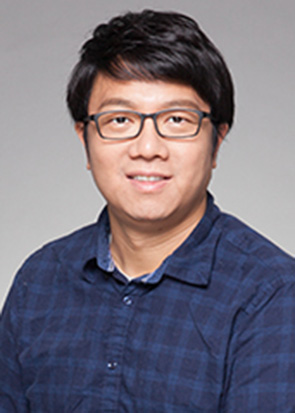}}]{Jun Wang} (Member, IEEE) received his Ph.D. degree in electrical engineering from Columbia University, New York, NY, USA, in 2011. Currently, he is a Professor at the School of Computer Science and Technology, East China Normal University, and an adjunct faculty member of Columbia University. From 2010 to 2014, he was a Research Staff Member at IBM T. J. Watson Research Center, Yorktown Heights, NY, USA. His research interests include machine learning, data mining, mobile intelligence, and computer vision. Dr. Wang has been the recipient of the Thousand Talents Plan in 2014.
\end{IEEEbiography}
}

\end{document}